\pdfoutput=1

\documentclass[11pt]{article}

\usepackage[final]{acl}

\usepackage{times}
\usepackage{latexsym}

\usepackage[T1]{fontenc}

\usepackage[utf8]{inputenc}

\usepackage{microtype}

\usepackage{inconsolata}

\usepackage{graphicx}
\usepackage{subcaption} 
\usepackage[most]{tcolorbox}
\usepackage{algorithm}
\usepackage{algorithmic}
\usepackage{booktabs}       
\usepackage{amsmath}        
\usepackage{tabularx}       
\usepackage{multirow}
\usepackage{tikz}
\usepackage{pgfplots}
\usepackage{makecell}
\usepackage{pgfplotstable}
\usepackage{filecontents}
\usepackage{siunitx}
\usepackage{array}

\usepackage{xcolor}         
\usepackage[dvipsnames]{xcolor}
\definecolor{c1}{HTML}{63b2ee}
\definecolor{c2}{HTML}{F27970}
\definecolor{c3}{HTML}{f8ac8c} 
\definecolor{c4}{HTML}{BEB8DC}

\definecolor{c5}{HTML}{ff7500}
\definecolor{c6}{HTML}{1280B0}
\definecolor{c7}{HTML}{17A1DE}
\definecolor{lightblue}{HTML}{53D2FF} 
\definecolor{lightyellow}{HTML}{F9DE00} 

\title{Bridging External and Parametric Knowledge: Mitigating Hallucination of LLMs with Shared-Private Semantic Synergy in Dual-Stream Knowledge}

\author{
 \textbf{Yi Sui\textsuperscript{1}},
 \textbf{Chaozhuo Li\textsuperscript{2}},
 \textbf{Chen Zhang\textsuperscript{3}},
 \textbf{Dawei Song\textsuperscript{1,4} \thanks{Corresponding author.}},
 \textbf{Qiuchi Li\textsuperscript{1}}
\\
 \textsuperscript{1}Beijing Institute of Technology, China \\
 \textsuperscript{2}Beijing University of Posts and Telecommunications, China \\
  \textsuperscript{3}Meituan, China
 \textsuperscript{4}The Open University, UK
 \\
 \small{
   \textbf{Correspondence:} \href{mailto:email@domain}{dawei.song2010@gmail.com}
 }
}

\begin{document}
\maketitle
\begin{abstract}
Retrieval-augmented generation (RAG) aims to mitigate the hallucination of Large Language Models (LLMs) by retrieving and incorporating relevant external knowledge into the generation process. However, the external knowledge may contain noise and conflict with the parametric knowledge of LLMs, leading to degraded performance. Current LLMs lack inherent mechanisms for resolving such conflicts. 
To fill this gap, we propose a Dual-Stream Knowledge-Augmented Framework for Shared-Private Semantic Synergy (DSSP-RAG). Central to it is the refinement of the traditional self-attention into a mixed-attention that distinguishes shared and private semantics for a controlled knowledge integration. An unsupervised hallucination detection method that captures the LLMs’ intrinsic cognitive uncertainty ensures that external knowledge is introduced only when necessary. To reduce noise in external knowledge, an Energy Quotient (EQ), defined by attention difference matrices between task-aligned and task-misaligned layers, is proposed. 
 Extensive experiments show that DSSP-RAG  achieves a superior performance over strong baselines.

\end{abstract}

\section{Introduction}


Large Language Models (LLMs) excel in natural language processing (NLP) tasks but face challenges in adapting to the rapid evolution of knowledge~\cite{meng2022mass, fan2025research}. The discrepancy between static pre-trained knowledge and continuously evolving external information leads to outdated, inconsistent, and insufficient knowledge within LLMs, contributing to hallucination and ultimately limiting model performance~\cite{huang2024survey, tonmoy2024comprehensive}.


\begin{figure*}
  \includegraphics[width=\textwidth]{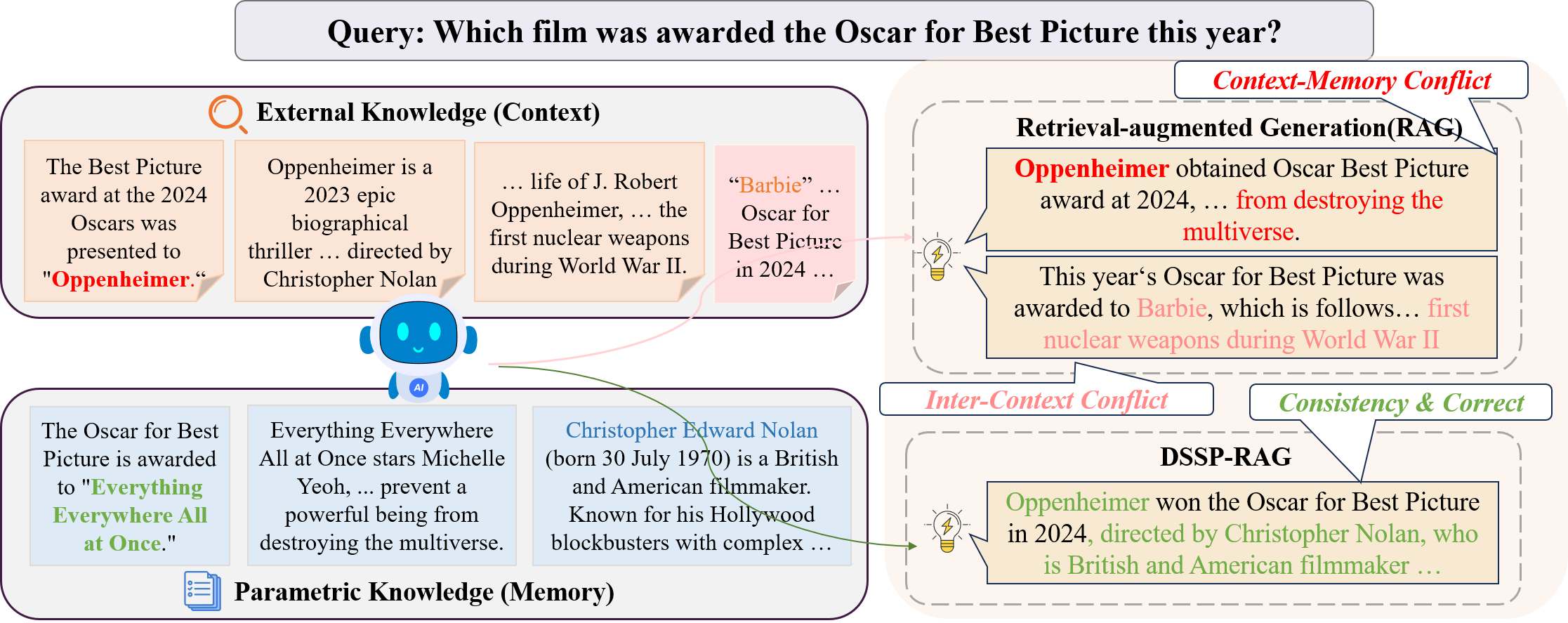}
  \caption{Comparison of Inference Results Using External vs. Parameterized Knowledge Across Methods}
  \label{fig:MethodCompare}
  \vspace{-15pt}
\end{figure*}

Retrieval-Augmented Generation (RAG) offers a cost-effective solution by incorporating retrieved external knowledge into the generation process of LLMs to alleviate hallucinations~\cite{ram2023context,qian2024memorag,yu2025rankrag, li2025loki}. However, as shown in Figure \ref{fig:MethodCompare}, the classical RAG methods face two core challenges. Firstly, there is a lack of an effective mechanism for real-time hallucination detection, resulting in blind injection of external knowledge into LLM~\cite{du2022syntheticdisinformationattacksautomated, pan-etal-2023-attacking}, but neglecting the need for proper integration of knowledge. Studies have shown that LLMs often exhibit a tendency to over-rely on external evidence~\cite{wu2024clasheval}, in which noise information may exist and inadvertently degrade the overall performance (inter-context conflict). Secondly, the external knowledge may conflict with the parametric knowledge acquired during LLM pre-training (context-memory conflict)~\cite{xu2024knowledge}. 
The lack of dedicated mechanisms to resolve such knowledge conflicts, coupled with the implicit knowledge encoding in current LLMs, would affect generation stability.


To ensure more effective RAG, hallucination detection and knowledge filtering are essential for identifying and mitigating irrelevant or misleading information prior to knowledge integration. Existing approaches of hallucination detection~\cite{xiong2023can, mahaut2024factual} and knowledge filtering~\cite{yu2023improving, wang2024blendfilter} often rely on supervised training that inherently limits the generalization to out-of-domain scenarios. Moreover, some denoising strategies use the confidence scores of the original LLM, rendering them susceptible to the model's internal biases~\cite{li2022house, zhou2024trustworthiness}.  Therefore, it is necessary to incorporate accurate and unsupervised detection modules.


More recent approaches to knowledge augmentation employ knowledge editing~\cite{zhang2024comprehensive, hase2024does, wang2025two} or iterative reasoning~\cite{wang2024knowledge}, but remain constrained by delayed integration and the absence of explicit semantic disentanglement~\cite{ju2024largelanguagemodelsencode}. The core limitation of current LLM architectures lies in their reliance on implicit fusion of internal and external knowledge, which not only lacks interpretability but also incurs substantial computational overhead due to extended context lengths~\cite{jin2024longcontextllmsmeetrag} and quadratic cost of the self-attention mechanism~\cite{yang2024gatedlinearattentiontransformers}.

To address the aforementioned challenges, we propose a Dual-Stream Knowledge-Augmented Framework for Shared-Private Semantic Synergy (DSSP-RAG). 
To tackle the inter-context conflicts, DSSP-RAG preemptively incorporates a two-stage control mechanism into RAG,  consisting of an unsupervised hallucination detection module to determine when the external knowledge is necessary, and a knowledge filtering module to maintain the accuracy of the external knowledge by filtering the noisy or redundant information. 
The former is grounded in intrinsic cognitive uncertainty of LLMs and detects hallucination by analyzing subspace variations in response to semantically equivalent prompts.
For the latter, we identify key and offset layer that differentially capture the task-aligned versus misaligned information, and propose an Energy Quotient (EQ) that is derived from the attention difference matrices between the pair of the key and offset layer to amplify key knowledge fragments while suppressing noise.


To mitigate the context-memory conflicts and optimize knowledge augmentation by maximizing the synergy between the model's intrinsic capabilities and external resources, we propose a mixed attention mechanism to decompose dual-stream knowledge into shared and private semantics in a pluggable way. 
Shared semantics captures the consistency between internal-external knowledge, aiming to improve the reliability of generation. On the other hand, private semantics preserves their unique contributions that are dynamically adjusted through a weighted fusion layer based on contextual relevance, thus mitigating the conflicts and improving generation accuracy. 

Our contributions are summarized as follows: (1) We propose an integrated framework that combines unsupervised hallucination detection, knowledge filtering and augmentation to precisely regulate the integration of external knowledge and mitigate hallucinations. (2) DSSP-RAG employs a mixed attention mechanism to resolve context-memory conflicts and enhance dual-stream knowledge utilization, improving generation reliability and accuracy. (3) Extensive experiments over four benchmarks show that our approach consistently outperforms the state-of-the-art baselines.

\begin{figure*}
  \centering
  \includegraphics[width=1\textwidth]{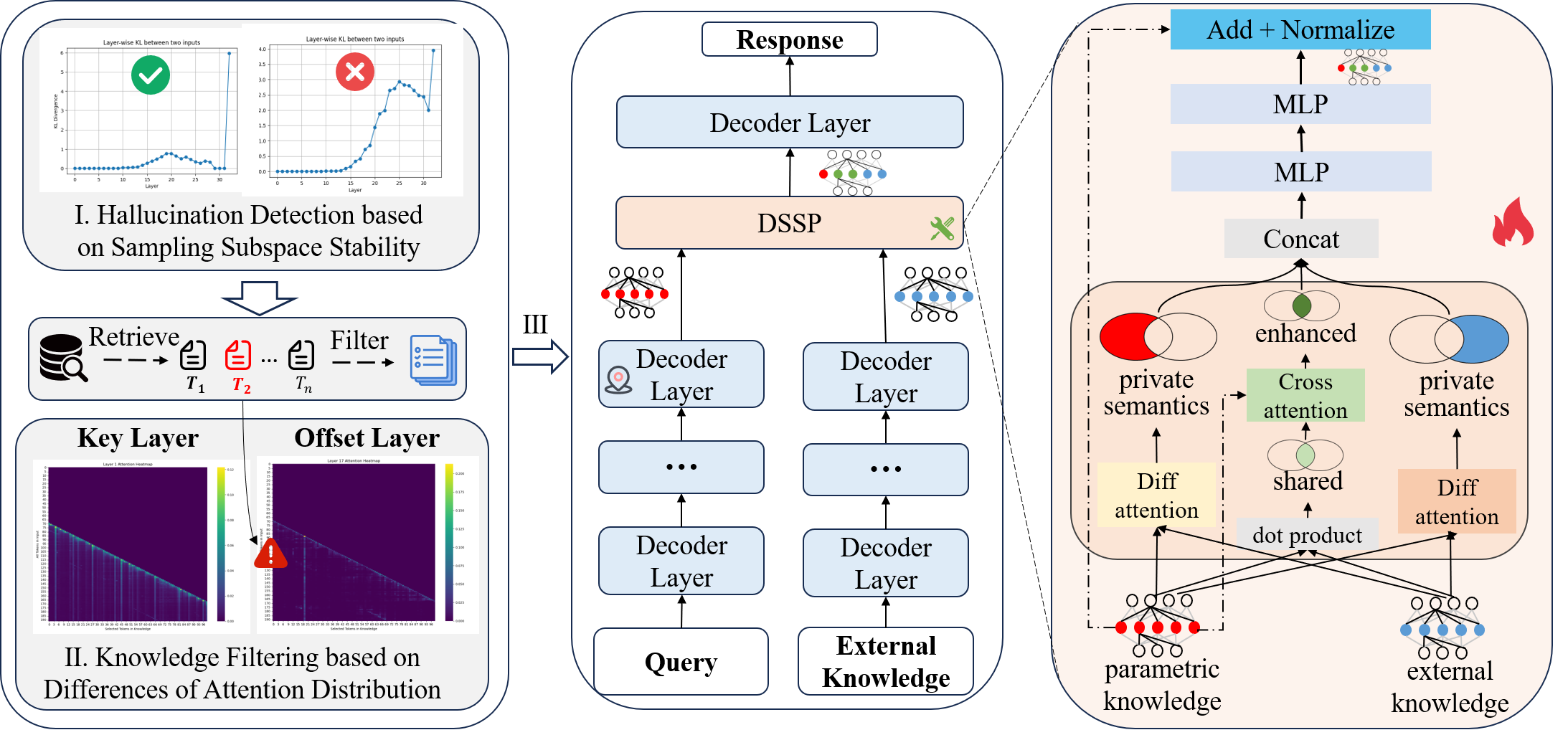}
  \caption{Dual-Stream Knowledge-Augmented Framework for Shared-Private Semantic Synergy (DSSP-RAG)}
  \label{fig:Framework}
  \vspace{-15pt}
\end{figure*}

\section{Related Work}

\textbf{Hallucination Detection.} 
LLMs acquire knowledge during pretraining, but the limitations of static corpora may result in factual inaccuracies, resulting in outputs based on unreliable or fabricated information that undermine response credibility~\cite{li2023copy, wang2023survey, he2024can}.
Researchers have proposed various techniques to detect hallucination. Current approaches primarily include training multi-layered probes to extract factual confidence scores from hidden states~\cite{azaria2023internal,kadavath2022language,burns2022discovering}, leveraging the averaged log probabilities assigned to a sequence of output tokens to estimate factual confidence~\cite{xiong2023can, yin2023large, lin2022teaching}, prompting the model to generate responses accompanied by confidence levels (Verbalization) ~\cite{tian2023just}, using prompts to guide the model in outputting specific tokens to assess factual accuracy, and evaluating consistency based on multiple sampling results~\cite{wang2022self, manakul2023selfcheckgpt, ding2024retrieve}. Recent studies indicate that training probes based on the model's internal states represents one of the most reliable methods for hallucination detection~\cite{mahaut2024factual}. Hallucination detection serves as the foundation for mitigating hallucination issues, and our proposed method stands out by eliminating the need for supervised training, ensuring greater generalizability.

\textbf{RAG for Hallucination.}
A straightforward approach to mitigate hallucination is to employ RAG methods to retrieve accurate and comprehensive external facts as a complementary resource to guide the generation process of LLMs~\cite{huang2023survey, gao2025synergizingragreasoningsystematic, wang2025chainofretrievalaugmentedgeneration}. 
However, in the face of complex or multi-hop reasoning tasks, the knowledge obtained through single-round retrieval often proves insufficient to provide adequate information for subsequent inference steps. Thus, ITER-RETGEN uses an iterative approach to coordinate retrieval and generation, leveraging LLM responses at each iteration to retrieve more relevant knowledge, thereby improving inference accuracy in subsequent rounds~\cite{shao2023enhancing, lee2025rearagknowledgeguidedreasoningenhances}. Moreover, Adaptive-RAG dynamically adjusts retrieval strategies based on task complexity, allowing flexible switching between single retrieval, iterative retrieval and no retrieval, to meet various task demands~\cite{jeong2024adaptive, guan2025deepragthinkingretrievalstep}. 
Integrating external knowledge through in-context learning can conflict with the model's internal knowledge and introduce noise, degrading LLM performance. We address this by filtering noise, resolving conflicts, and enabling collaborative utilization of dual-stream knowledge to mitigate hallucinations.

\section{Methodology}
\label{method}

\subsection{Overview}
 
Figure~\ref{fig:Framework} presents an overview of DSSP-RAG, which comprises three core modules designed to integrate external knowledge with internal representations while mitigating hallucinations. First, DSSP-RAG identifies hallucination-prone instances through subspace stability analysis, enabling adaptive retrieval timing. Second, an attention difference matrix is constructed to filter noise from retrieved knowledge, ensuring the relevance of the incorporated information. Finally, the DSSP module proposes a mixed attention mechanism to disentangle external and internal knowledge into shared and private semantics, resolving potential conflicts and enhancing the reliability of the output. 

\subsection{Hallucination Detection Based on Stability of Sampling Subspaces}


Hallucination detection provides a guiding signal for the necessity and timing of integrating external knowledge. To overcome domain-specific limitations and the reliance on labeled data in previous work~\cite{azaria2023internal, kuhn2023semantic}, we propose an unsupervised detection method based on the so-called cognitive uncertainty~\cite{yadkori2024believe}, which manifests as inconsistent outputs in response to semantically equivalent but syntactically varied prompts. 
However, the outputs can be influenced by input noise and model biases.
LLMs reason within high-dimensional sampling (probabilistic) subspaces shaped by parametric knowledge from pre-training. 
Thus, we employ the stability of the model's sampling subspaces~\cite{ji2024llm, zhang2024truthx} as a metric to detect hallucinations, enabling an improved generalization across tasks and a more reliable assessment of the reasoning process.

To this end, we prompt the LLM to automatically generate syntactically distinct but semantically similar alternative expressions for each query. 
To assess the stability of sampling subspaces, we employ Jensen-Shannon Divergence (JSD) to quantify the differences between the subspaces of paired queries $(x, \hat{x})$ across model layers. 
\vspace{-3pt}
\begin{equation}
    d(s_l (\cdot |x),\hat{s _l }(\cdot | \hat{x} ))=\mathrm{JSD}(s_l (\cdot |x)||\hat{s _l }(\cdot | \hat{x} )) \text{.}
    \vspace{-3pt}
\end{equation}

Here, $s_l$ denotes the sampling subspace activated by the LLM in layer $l$ for a given input.
The JSD distance $d$ below a predefined threshold $\delta$ suggests an effective internalization of correct knowledge from the pretraining corpus of the LLM.
Cognitive uncertainty may cause a divergence in response, leading to an increase in JSD in deeper layers, as illustrated in the top-left subfigure of Figure~\ref{fig:Framework}. A $d$ exceeding the threshold would indicate an instability in the sampling subspace, leading to an increased risk of hallucination. 
JSD is preferred over alternatives like KL-divergence due to its symmetric and bounded properties, see Appendix~\ref{sec:halluAppend} for a detailed analysis.


\subsection{Filtering of External Knowledge}

\begin{figure}
    \centering
    \includegraphics[width=0.75\linewidth]{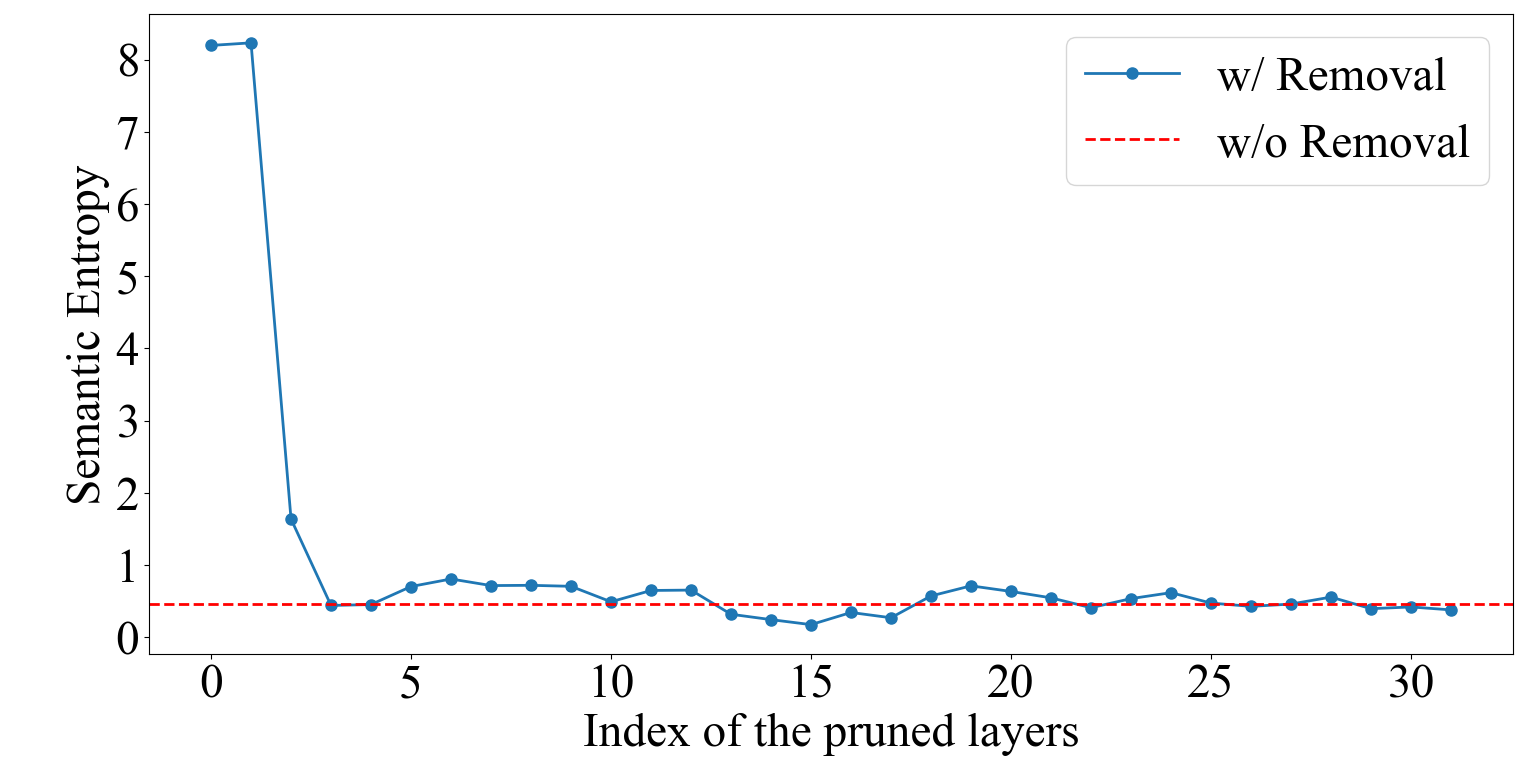}
    \caption{Performance variation with layer-wise pruning}
    \label{fig:earlyStop}
    \vspace{-15pt}
\end{figure}

The core of the RAG is to enhance the LLM performance by retrieving external factual knowledge to support generation process. Current knowledge filtering approaches exhibit poor domain generalization and heightened sensitivity to model-induced biases~\cite{fan2024survey}.
Therefore, inspired by the work of ~\citet{zhang2023crashclusteringremovingsharing}, we perform a layer-wise pruning study to assess the influence of external knowledge on LLM's decision-making.

By systematically removing each layer of the LLM (denoted as w/ Removal) and comparing the performance differences with the original model (denoted as w/o Removal), the variations are quantitatively evaluated using semantic entropy, which measures the uncertainty of the model's output:
{
\setlength{\abovedisplayskip}{8pt}  
\setlength{\belowdisplayskip}{4pt}  
\begin{equation}
H(Y|Q) = -\sum_{y} p(y|Q) \, \mathrm{log} \, p(y|Q) \text{.}
\end{equation}
}
Here, $p(y|Q)$ denotes the model's predicted probability for a given output token $y$, conditioned on input query $Q$. The higher entropy value indicates a greater uncertainty in the model prediction.


The results (Figure~\ref{fig:earlyStop}) reveal performance fluctuations caused by layer ablation. Specifically, removing certain layers leads to an increased semantic entropy, underscoring their role in encoding task-relevant information. These layers often exhibit broadly distributed attention to contextual semantics. In contrast, eliminating other layers reduces entropy, suggesting that excessively narrow attention may encode noise or irrelevant features.
Thus, we propose a classification of layers:

\vspace{-0.5em}
\begin{itemize}
    \item Key Layer whose removal causes significant performance degradation, indicating its role in extracting task-relevant semantics and constructing global representations aligned with inference objectives;
    \vspace{-0.6em}
    \item Offset Layer whose removal leads to the greatest performance improvement, suggesting it amplifies noise or irrelevant features due to attention misalignment with task goals.
    
\end{itemize}
\vspace{-0.5em}

Furthermore, from an information-theoretic perspective, we construct the Energy Quotient (EQ) as a filtering matrix by analyzing the difference in attention distribution, \(\Delta A = A_{\alpha} - A_{\beta}\), between an offset layer (\(\alpha\)) and a key layer (\(\beta\)), thereby quantifying the significance of external context:
\vspace{-4pt}
\begin{equation}
    EQ_{i}=\frac{\exp(-\lambda \Delta A_{i})}{\sum_{j} \exp(-\lambda \Delta A_{j})} \text{.}
\end{equation}
Here, $\lambda$ denotes the temperature coefficient. The higher $\Delta A_{i}$, the poorer the stability of the representation space. Thus, a smaller $EQ_{i}$ indicates that the feature $i$ increases uncertainty in the reasoning process and is more likely to be identified as noise. 
In contrast, a larger $EQ_i$ signifies a higher informational relevance and a greater contribution to the reasoning process.

Considering that the degree of noise varies in different tasks, we further introduce a dynamic weighting coefficient $\varepsilon $, formulated as follows, to adjust the filtering effect of EQ on the external context. 
\vspace{-14pt}

\begin{equation}
\varepsilon = 
\begin{cases}
\log\left(1 - \frac{\Delta \hbar}{\hbar_{\text{orig}}} \right), & \Delta \hbar < -0.1 \\
0, & \text{otherwise}
\end{cases}
\label{eq:entropy_penalty}
 \vspace{-5pt}
\end{equation}

Here, $\bigtriangleup \hbar = \hbar_{\alpha}-\hbar_{orig}$, where $\hbar_{orig}$ represents the semantic entropy of the original model response with external context, and $\hbar_{\alpha}$ corresponds to the semantic entropy after removing the offset layers. A negative and larger absolute value of $\bigtriangleup \hbar$ indicates that the information attended to by the offset layers has a significant negative impact, representing increased noise. Conversely, if the difference is close to zero or positive, it indicates low noise in external information, and knowledge filtering is not required. Finally, the encoded external information $\hat{D}$ is updated based on EQ-based filtering and coefficient weighting, as follows:
\vspace{-4pt}
\begin{equation}
\hat{D} = 
\begin{cases}
\varepsilon \cdot EQ \cdot D, & \Delta \hbar < -0.1 \\
D, & \text{otherwise}
\end{cases}
\label{eq:filtered_knowledge}
\end{equation}

\subsection{Dual-Stream Knowledge Augmentation}


A primary limitation for LLMs in resolving context-memory conflicts stems from the absence of explicit semantic decomposition in the conventional attention, hindering the distinction between private (source-specific) and shared (overlapping) semantics~\cite{xie2024adaptivechameleonstubbornsloth, li2024generalizing, Dong2025Decouplingknowledgeandcontext}. The inherent uniformity bias in attention weights further exacerbates this issue by amplifying the redundancy in shared semantics while neglecting the conflicts in private semantics, leading to ambiguous output. Furthermore, the extended context length introduced by external knowledge increases computational overhead and diminishes the model's focus on salient information~\cite{jin2024longcontextllmsmeetrag, liu-etal-2024-lost}.
In response to these challenges, a mixed attention of the DSSP module is proposed to decompose dual-stream knowledge into shared and private semantics at the target layer, identified via the maximal JSD between semantically identical prompts. DSSP dynamically balances the integration of parametric knowledge and external sources while mitigating the context-memory conflicts. A mathematical derivation of the mixed attention is detailed in Appendix~\ref{sec:mathDsspAppend}.

Shared semantics refers to the overlapping portion of knowledge between internal knowledge $I$ and external knowledge $\hat{D}$, which exhibits high consistency between different sources of knowledge. To capture this shared information, we compute the similarity matrix between dual-stream knowledge using the dot product similarity:
\vspace{-5pt}
\begin{equation}
\begin{split}   
sim(I,\hat{D})=\varphi (\frac{ (W_{share} I)(W_{share} \hat{D})}{\sqrt{d_{k}}}) \text{,}
\end{split}
\end{equation}
where $\varphi$ denotes the softmax function. Based on the similarity matrix, the top-$T$ tokens from external knowledge most similar to internal knowledge are selected as shared semantics, denoted as $U_{\text{share}}$. Then a cross-attention mechanism is applied to derive $U_{\text{enhance}}$, which integrates $U_{\text{share}}$ into the internal knowledge stream, amplifying the consistency between external and internal representations.

Private semantics refers to the unique information within internal $I$ or external knowledge $\hat{D}$, which helps fill the gaps in the other source. This provides additional useful information for the model's predictive behavior.
To extract these private features, we introduce a differential attention mechanism that subtracts the shared components from each knowledge stream. Self-attention within each source models token-level dependencies and contextual interactions.
\vspace{-14pt}

\begin{equation}
    \tau (X, X)=\varphi(\frac{(W_{Q}^{s} X)(W_{K}^{s} X)}{\sqrt{d_{k}}})(W_{V}^{s} X) \text{.}
\end{equation}
Meanwhile, the cross-attention models the mutual influence and semantic relationships between the dual knowledge, capturing the shared information:
\vspace{-27pt}

\begin{equation}
    \eta (X, Y)=
    \varphi(\frac{(W_{Q}^{c} X)(W_{K}^{c} Y)}{\sqrt{d_{k}}})(W_{V}^{c} Y) \text{.}
\end{equation}

Thus, we define the differential attention mechanism as the difference between self-attention $\tau (\cdot )$ of single knowledge source and cross-attention $\eta (\cdot )$ of dual-stream knowledge:
\vspace{-3pt}
\begin{equation}
    U_{\text{private}} = \mathcal{D}_{\text{attn}}(X, Y) = \tau(X, X) - \eta(X, Y) \text{.}
\end{equation}
When $X$ and $Y$ represent internal knowledge and external knowledge respectively, the private semantics within the internal knowledge can be extracted through differential attention, and similarly, the private semantics within the external knowledge can be obtained. The explicit decoupling mechanism enables DSSP-RAG to adaptively reweight knowledge sources according to task demands:

\vspace{-16pt}
\begin{equation}
    U = \mathrm{concat}(U_{enhance},U_{private}^{I},U_{private}^{\hat{D}}) \text{,}
\end{equation}
\begin{equation}
    \begin{split}
    \hat{U} = \operatorname{LN}(W_{o}\left(\operatorname{ReLU}\left(W_{f}U+b_{f}\right)\right)+b_{o}) \text{.}
    \end{split}
\end{equation}
To ensure the stability of model training, we apply a residual connection to combine the original representation $I$ and the aggregated knowledge representation $\hat{U}$. This combined representation is then passed into the subsequent layers of the model encoding and further-depth knowledge fusion.

\subsection{Loss Function}
LLMs are typically trained with cross-entropy loss, but their inherent confidence bias may cause overconfidence in incorrect predictions or undue uncertainty in correct ones. Moreover, models without knowledge adaptation struggle to effectively attribute the influence of external knowledge , resulting in abrupt probability distribution shifts that may undermine the predictive accuracy. 

To address this issue, we introduce a conditional entropy $H(P(\hat{U}|I))=- {\textstyle \sum_{\hat{y}}P(\hat{y}|I)logP(\hat{y}|\hat{U})}$ to regulate the predictive uncertainty, allowing the model to dynamically adjust confidence based on the reliability of external knowledge. Accurate external knowledge reduces uncertainty, while noisy knowledge increases it, providing a basis for filtering external information. However, relying solely on conditional entropy may compromise training stability, as it lacks constraints on distribution shifts across different inputs, potentially leading to overfitting or bias toward erroneous knowledge.
To enhance stability, we incorporate the KL divergence $D_{KL}(P(\hat{y}|\hat{U})||P(\hat{y}|I))$ to regularize the distributional shift of external knowledge, preventing excessive deviations. This regularization ensures that the model maintains prediction stability while avoiding overreliance on noisy information.
The final loss function is then formulated as:
\vspace{-5pt}
\begin{equation}
    L=L_{CE}+\mu H(P(\hat{U}|I))+\nu D_{KL}   \text{,}
    \vspace{-5pt}
\end{equation}
where $L_{CE}$ represents the cross-entropy loss, $\mu$ and $\nu$ respectively controls the regularization strength of conditional entropy and KL divergence.

\begin{table*}[h!]
\centering
\small
\setlength{\tabcolsep}{5.5pt}
\begin{tabular}{c|c|c|c|c|c|c|c|c|c|c}
\toprule
\textbf{Model} & \multicolumn{5}{c|}{\textbf{LLaMA2-7B-Chat}} & \multicolumn{5}{c}{\textbf{LLaMA3-8B-Instruct}} \\
\midrule
\multirow{2}*{\textbf{Method}}  & 
\textbf{NQ} & \textbf{TriviaQA} & \textbf{HotpotQA} & \textbf{2Wiki} &
\textbf{PubH} &
\textbf{NQ} & \textbf{TriviaQA} & \textbf{HotpotQA} & \textbf{2Wiki} &
\textbf{PubH}  \\ 
\cmidrule(r){2-2} \cmidrule(r){3-3} \cmidrule(r){4-4} \cmidrule(r){5-5} \cmidrule(r){6-6} \cmidrule(r){7-7} \cmidrule(r){8-8} \cmidrule(r){9-9} \cmidrule(r){10-10} \cmidrule(r){11-11}
 
  & \textbf{EM} & \textbf{EM}  & \textbf{F1} & \textbf{F1} & \textbf{ACC} & \textbf{EM} & \textbf{EM} & \textbf{F1} & \textbf{F1} & \textbf{ACC} \\
 
\midrule
\multicolumn{11}{c}{\textit{Simple RAG}} \\
 No Retrieval & 13.8 & 30.5 &  27.5  & 22.3 & 33.4
    & 22.6 & 55.7  & 28.4  & 33.9 & 50.2 \\
 SR-RAG & 20.7 & 42.5 & 25.0  & 25.5 & 30.7 
    & 30.1 & 58.3 & 35.8 & 29.6 & 51.3\\
  \midrule
 \textbf{DSSP\_SR} & 19.5 & 45.3  & 29.3 & 26.7 & 34.2
    & 28.6 & 58.9 & 37.6  & 36.8 & 52.7\\
\midrule
 \multicolumn{11}{c}{\textit{Adaptive RAG}} \\
 Self-RAG & \textbf{32.3} & 57.0 & 17.5 & 19.6 & - 
    & \textbf{36.4} & 38.2 & 29.6 & 25.1 & -\\
 FLARE & 25.3 & 50.7 & 22.1 & 24.3 & 37.1
    & 22.5 & 55.8 & 28.8  & 33.9 & 44.5\\
 DRAGIN & 23.2 & 54.0 & 29.2 & 30.0 & 52.4
    & - & - & 44.6 & 37.8 & 66.6\\
 SEAKR  & 25.6 & 54.1 & \textbf{38.1} & 36.0 & 59.3  
    & 31.0 & 59.4 & 47.7 & 48.1 & 70.2\\
\midrule
 \textbf{DSSP\_AR} & 25.1 & $\textbf{55.4}^{\dag}$ & 37.7 & $\textbf{37.8}^{\dag}$ & $\textbf{62.1}^{\dag}$
    & 30.8 & $\textbf{60.2}^{\dag}$ & $\textbf{48.6}^{\dag}$  & $\textbf{49.8}^{\dag}$ & $\textbf{73.5}^{\dag}$ \\

\textbf{$\Delta$ (\%)} & $1.95^{\downarrow}$ & $\textbf{2.40}^{\uparrow}$ & $1.05^{\downarrow}$ & $\textbf{5.00}^{\uparrow}$  & $\textbf{4.72}^{\uparrow}$  
    & $0.65^{\downarrow}$ & $\textbf{1.35}^{\uparrow}$ & $\textbf{1.89}^{\uparrow}$  & $\textbf{3.54}^{\uparrow}$ & $\textbf{4.55}^{\uparrow}$  \\

\bottomrule
\end{tabular}
\caption{Performance comparison across methods and datasets on Llama family. $\dag$ refers to a statistically significant improvement achieved by DSSP-RAG over SEAKR (the best performing baseline) according to paired t-test ($p<0.05$). Self-RAG is fine-tuned from LLaMA2-7B-chat using NQ-style data.}
\vspace{-15pt}
\label{tab:llamaTable}
\end{table*}

\section{Experiments}
\label{exper}

\subsection{Tasks and datasets}
DSSP-RAG is evaluated on five datasets, including two single-hop QA datasets: Natural Questions (NQ)~\cite{kwiatkowski2019natural} which is based on real user queries, and TriviaQA~\cite{joshi2017triviaqa} that consists of trivia questions; and two multi-hop QA datasets: Adversarial HotpotQA~\cite{yang2018hotpotqa}, and 2WikiMultihopQA~\cite{ho2020constructing}, which assess cross-document reasoning and information integration using Wikipedia data. In addition, PubHealth~\cite{kotonya2020explainableautomatedfactcheckingpublic} is a closed-set generative dataset that comprises medical claims on various biomedical topics. More details are included in the Appendix~\ref{sec:dataAppend}.


\subsection{Experimental setups and baselines}
All experiments are conducted on NVIDIA RTX A6000 GPU with 49 GB of memory.
The training data consists of question-answer pairs from HotpotQA and 2WikiMultihopQA, leveraging the supporting and non-supporting facts provided within the datasets. The effectiveness of DSSP-RAG is validated in different model architectures, including \textbf{LLaMA2-7B-Chat}~\cite{touvron2023llama}, \textbf{LLaMA3-8B-Instruct}~\cite{meta2024introducing} and \textbf{Qwen2.5-3B-Instruct}~\cite{qwen2025qwen25technicalreport}.  In our main experiment, we employ the BM25~\cite{robertson2009probabilistic} as retriever, utilizing the English version of Wikipedia dumped on December 20, 2018 as the retrieval corpus and retrieving five documents.


To evaluate the effectiveness of DSSP-RAG, we compare it with two representative baselines of RAG.
(1) \textbf{Simple RAG}(SR-RAG): which directly prepends the external knowledge obtained from a single retrieval to the LLM's prompt.
(2)\textbf{Adaptive RAG}: an active RAG framework that decides when and what to retrieve during generation, including \textbf{Self-RAG}~\cite{asai2023self}, \textbf{FLARE}~\cite{jiang2023active}, \textbf{DRAGIN}~\cite{su2024dragin}, \textbf{SEAKR}~\cite{yao2024seakr}. 
Following SEAKR, we re-implement FLARE with IRCoT strategy~\cite{trivedi2022interleaving} to support evaluation on complex QA. IRCoT interweaves CoT reasoning with retrieval-augmented generation strategy. 
Meanwhile, we implement DSSP-RAG with simple and adaptive retrieval, yielding two variants: \textbf{DSSP\_SR} and \textbf{DSSP\_AR}.

\subsection{Results and analysis}
\subsubsection{Main results}

The experimental results in Table~\ref{tab:llamaTable} demonstrate that DSSP-RAG consistently achieves optimal performance in different retrieval strategies over baseline methods.
For complex queries, the static and broad nature of simple retrieval strategies often fails to provide precise matches, thus impairing model performance. However, 
DSSP-RAG effectively leverages both internal and external knowledge while mitigating noise interference, leading to significant performance gains.

\begin{table}[h!]
\centering
\small
\begin{tabular}{l|c|c|c|c}
\toprule
\multirow{2}{*}{\textbf{Method}} & \textbf{NQ} & \textbf{TriviaQA} & \textbf{HotpotQA} & \textbf{2Wiki}  \\ 
\cline{2-2} \cline{3-3} \cline{4-4} \cline{5-5}
\noalign{\vskip 1mm}
& \textbf{EM} & \textbf{EM} &  \textbf{F1} & \textbf{F1}  \\ 
\midrule
\multicolumn{5}{c}{\textit{Simple RAG}} \\
No Retrieval & 29.3 &  57.6 & 28.7 & 30.2 \\
SR-RAG & 38.5  & 60.8 & 37.1 & 27.4  \\
\midrule
DSSP-SR & 39.7 & 61.3 & 40.1 & 32.3  \\
\midrule
\multicolumn{5}{c}{\textit{Adaptive RAG}} \\
FLARE  & 30.1 & 58.8 & 27.5 & 31.7 \\
DRAGIN & - & - & 42.6 & 38.7  \\
SEAKR & 42.1  & 63.7 & 48.1 & 40.1  \\
\midrule
\textbf{DSSP\_AR} & \textbf{43.6} & \textbf{65.3} & \textbf{48.4} & \textbf{44.2} \\
\textbf{$\Delta$ (\%)}  & $\textbf{3.56}^{\uparrow}$ & $\textbf{2.51}^{\uparrow}$ & $\textbf{0.62}^{\uparrow}$ & $\textbf{10.22}^{\uparrow}$ \\  
\bottomrule
\end{tabular}
\caption{Performance comparison across different methods and datasets on Qwen2.5-3B-Instruct.}
\label{tab:Qwen}
\vspace{-20pt}
\end{table}

Although adaptive retrieval expands the knowledge scope, it also introduces more noise and redundancy. DSSP-RAG mitigates this by integrating a hallucination detection module to assess the necessity of external knowledge and leveraging attention difference matrices to filter irrelevant content. In contrast, baseline methods, except SEAKR, lack such mechanisms, resulting in interference with internal knowledge by noise and reduced output reliability.
DSSP-RAG slightly underperforms SEAKR on HotpotQA with LLaMA2-7B-Chat, as SEAKR’s self-aware signals better capture uncertainties in critical intermediate steps in multi-hop reasoning.
SEAKR utilizes LLM-based uncertainty to guide retrieval but fails to resolve conflicts between external and internal knowledge. In contrast, DSSP-RAG's mixed attention mechanism exploits shared semantics for reliable information activation and adaptively balances private semantics to mitigate conflicts and external interference. This adaptability ensures stable performance of DSSP\_AR in cross-domain tasks, whereas baseline models suffer from fluctuations due to domain shifts.

DSSP-RAG consistently outperforms baselines across tasks, with improvements varying by model scale and dataset characteristics. Gains are most significant in multi-hop QA, which requires extensive knowledge integration, while single-hop QA tasks like NQ show limited improvement. As illustrated in Table~\ref{tab:Qwen}, DSSP-RAG in Qwen2.5-3B-Instruct achieves higher performance gains than baselines compared to LLaMA3-8B-Instruct. These results across architectures and scales demonstrate the strong robustness of DSSP-RAG. In addition, we further evaluate the generalizability of DSSP-RAG on the ASQA (long-form QA) dataset, see Appendix~\ref{sec:ASQAResult} for detailed results.


\subsubsection{Evaluation of DSSP under Knowledge Conflict Scenarios}

To evaluate DSSP-RAG’s effectiveness in resolving knowledge conflicts, we compare it with two recent state-of-the-art baselines: EditCoT~\cite{wang2024knowledge} and CK-PLUG~\cite{bi2025parameters}. For a fair comparison, hallucination detection and knowledge filtering are removed from DSSP-RAG, leaving only the DSSP module that modulates reliance between parametric and retrieved knowledge. As shown in Table 4, DSSP consistently outperforms both baselines across multiple backbone models.

EditCoT updates knowledge at the chain-of-thought level, constraining activation of parametric knowledge, while CK-PLUG regulates reliance on parametric versus retrieved knowledge through entropy-based confidence gain, which is prone to output bias. In contrast, DSSP integrates knowledge at the representation level via mixed-attention, enabling fine-grained coordination between internal and external knowledge. These results highlight DSSP’s superior robustness and adaptability in resolving knowledge conflicts.
\vspace{-7.5pt}

\begin{table}[ht]
\centering
\small
\begin{tabular}{l| c c c}
\toprule
\multirow{2}{*}{\textbf{Method}} & \multirow{2}{*}{\textbf{Method}} & \textbf{TriviaQA} & \textbf{HotpotQA} \\
\cline{3-3} \cline{4-4}
\noalign{\vskip 1mm}
& & \textbf{EM} & \textbf{F1} \\
\midrule
\multirow{3}{*}{\textbf{\makecell{LLaMA3\\-8B-\\Instruct}}} 
 & EditCoT        & 54.7 & 43.1 \\
 & CK-PLUG & 57.5 & 45.8 \\
 & w/ DSSP    & \textbf{58.3} & \textbf{46.4} \\
\midrule
\multirow{3}{*}{\textbf{\makecell{Qwen2.5\\-3B-\\Instruct}}} 
 & EditCoT        & 59.2 & 44.6 \\
 & CK-PLUG & 61.8 & 46.1 \\
 & w/ DSSP    & \textbf{63.2} & \textbf{47.1} \\
\bottomrule
\end{tabular}
\caption{Performance comparison of knowledge conflict mitigation technologies.}
\label{tab:knowledge_conflict}
\vspace{-14pt}
\end{table}


To evaluate the effectiveness of DSSP module in balancing internal and external knowledge under conflicts, we follow CK-PLUG by constructing a synthetic RAG setup with two deliberately incorrect statements in the retrieved context, enabling stress-testing of LLM conflict resolution. Table~\ref{tab:dssp_performance} reports three metrics: ConR (alignment with the retrieved context), ParR (alignment with model parameters), and EM.

\vspace{-7.5pt}
\begin{table}[ht]
\centering
\small
\begin{tabular}{l| c c c c}
\toprule
\textbf{Model} & \textbf{Method} & \textbf{ConR} & \textbf{ParR} & \textbf{EM} \\
\midrule
\multirow{3}{*}{\textbf{\makecell{LLaMA2\\-7B-\\Chat}}} 
    & w/o RAG   & -     & -     & 27.5 \\
    & w/o DSSP  & 65.7  & 30.1  & 23.6 \\
    & w/ DSSP   & 51.6  & 32.3  & 25.9 \\
\midrule
\multirow{3}{*}{\textbf{\makecell{LLaMA3\\-8B-\\Instruct}}} 
    & w/o RAG   & -     & -     & 28.4 \\
    & w/o DSSP  & 56.8  & 25.4  & 29.7 \\
    & w/ DSSP   & 44.7  & 31.6  & 31.5 \\
\midrule
\multirow{3}{*}{\textbf{\makecell{Qwen2.5\\-3B-\\Instruct}}} 
    & w/o RAG   & -     & -     & 28.7 \\
    & w/o DSSP  & 46.1  & 21.7  & 32.1 \\
    & w/ DSSP   & 37.4  & 28.3  & 33.6 \\
\bottomrule
\end{tabular}

\caption{Performance (\%) of DSSP Module in controlling knowledge reliance.}
\label{tab:dssp_performance}
\vspace{-14pt}
\end{table}


Across all model configurations, replacing standard self-attention with our mixed-attention mechanism consistently reduces reliance on misleading context (lower ConR) while strengthening alignment with parametric knowledge (higher ParR). This shift is crucial when external knowledge introduces noise or contradictions. Furthermore, DSSP consistently improves EM over baselines, underscoring the effectiveness of mixed-attention in synergistically leveraging internal and external knowledge for enhanced reasoning and conflict resolution.
See Appendix~\ref{sec:caseStudy} for Representative examples.

\subsubsection{Ablation study}

We conduct an ablation study on Llama-7b-chat to assess the components of DSSP-RAG  (Table~\ref{ablation-table}). The DSSP module is critical, as its removal (w/o DSSP) causes a substantial drop in EM and F1 across datasets. By converting self-attention into mixed attention, DSSP enables effective integration of external knowledge while mitigating conflicts and improving the reliability of the output.

Knowledge filtering (K.F.) has a greater impact than hallucination detection (H.D.). Removing K.F. (w/o K.F.) leads to a more significant decline, especially in F1, underscoring its role in removing irrelevant or noisy knowledge and emphasizing the importance of external knowledge accuracy for response quality.
Although removing hallucination detection (w/o H.D.) results in a smaller performance decrease, it still noticeably degrades generation quality, indicating that unregulated incorporation of external knowledge can introduce inconsistencies or biases. The H.D. module mitigates this by identifying hallucination-prone cases.

\vspace{-7pt}
\begin{table}[h!]
\centering
\small
\setlength{\tabcolsep}{5.5pt}
\begin{tabular}{l|cc|cc|c}
\toprule
\multirow{2}{*}{\textbf{Method}} & \multicolumn{2}{c|}{\textbf{HotpotQA}} & \multicolumn{2}{c|}{\textbf{2Wiki}} & \textbf{TriviaQA} \\ 
\cline{2-3} \cline{4-5} \cline{6-6}
\noalign{\vskip 1mm}
& \textbf{EM} & \textbf{F1} & \textbf{EM} & \textbf{F1} & \textbf{EM} \\ 
\midrule
DSSP-RAG & 28.5 & 37.7 & 31.6 & 37.8 & 55.4 \\
\midrule
\multicolumn{6}{c}{\textit{Ablating hallucination detection module}} \\
Trained Probes & 28.6 & 37.1 & 32.1 & 37.7 & 54.5 \\
Verbalization & 27.1 & 36.6 & 29.8 & 36.5 & 52.7 \\
Hyb-Cons & 27.8 & 36.8 & 31.4 & 37.1 & 54.3 \\
\midrule
\multicolumn{6}{c}{\textit{Ablating knowledge filtering module}} \\
Entropy-based & 26.7 & 36.9 & 29.1 & 35.6 & 53.3 \\
Attention-based & 27.4 & 37.1 & 30.7 & 36.5 & 54.1 \\
\midrule
w/o H.D. & 27.6 & 36.8 & 30.1 & 36.0 & 52.9 \\
w/o K.F. & 27.0 & 37.0 & 29.5 & 35.6 & 53.1 \\

w/o DSSP & 25.3 & 34.8 & 26.9 & 33.7 & 51.7 \\
\bottomrule
\end{tabular}
\caption{Ablation study.}
\label{ablation-table}
\vspace{-10pt}
\end{table}

Furthermore, we replace the hallucination detection module in DSSP-RAG with two existing methods: Trained Probes, Verbalization and Hybrid Consistency (Hyb-Cons). Verbalization suffers from LLMs’ overconfidence~\cite{ni2024llms}, underperforming even the baseline w/o H.D., whereas our unsupervised detection based on sampling subspace stability matches the supervised Trained Probes in effectiveness.
The Hyb-Cons~\cite{ding2024retrieve} integrates cross-language and cross-model scores to assess the consistency of perturbed responses for identical queries. 
DSSP-RAG framework mitigates hallucination more robustly than consistency-based baselines by analyzing sampling subspace stability rather than surface-level output.  Its lightweight, unsupervised design integrates efficiently into RAG pipelines without auxiliary models or complex decoding, demonstrating both effectiveness and generalizability.

To better assess the effectiveness of the EQ-based knowledge filtering module, we replaced the module in DSSP-RAG with two representative approaches: Entropy-based Filtering~\cite{qiu2024entropy}, which assigns preference weights to documents based on token distribution entropy, and Attention-based Filtering~\cite{peng2025cafe}, which scores relevance via cross-layer attention with a fixed bias.
The results demonstrate that EQ-based method achieves superior factual accuracy and more robust performance across datasets. 
DSSP-RAG departs from raw attention-based importance measures by adopting an information-theoretic view, modeling cross-layer attention variation to capture semantic stability. A temperature-controlled Energy Quotient enhances sensitivity to representational instability, while an entropy-aware adaptive filter dynamically adjusts external knowledge contributions, thereby reducing irrelevant interference and improving robustness.

\subsubsection{Hyperparameter and efficiency analysis}

We evaluate DSSP-RAG and two baselines on 2WikiMultihopQA using LLaMA-7B-Chat with varying numbers of retrieved documents (Table~\ref{tab:knowNumber}). SR-RAG initially benefits from knowledge augmentation, but deteriorates with more than five documents due to increased noise. SEAKR, despite incorporating filtering and ranking, struggles with relevance in extended contexts. In contrast, DSSP-RAG consistently outperforms baselines by leveraging mixed attention to disentangle shared and private semantics, enabling effective knowledge integration and mitigation of hallucinations.

Furthermore, inference efficiency is measured with each method retrieving five external documents per query. The results show that DSSP-RAG effectively balances inference efficiency and resource consumption. Due to the design of DSSP module, its autoregressive decoding time is 0.72s substantially faster than SR-RAG (1.6s) and SEAKR (1.25s), which encodes internal and external knowledge in parallel and delays fusion via hybrid attention until the target layer. Despite introducing moderate overhead, DSSP-RAG maintains a low peak memory usage (17.34), outperforming SEAKR (24.03) and remaining comparable to SR-RAG (16.08). SEAKR's higher resource demand stems from its reliance on repeated LLM sampling.

\begin{table}
    \centering
    \small
    \setlength{\tabcolsep}{5pt}
    \begin{tabular}{l| c c c c c c c}
    \toprule
    \textbf{Method} & \textbf{3} & \textbf{5} & \textbf{10} & \textbf{20} & 
    \textbf{Inf.} & \textbf{Mem.} \\
    \midrule
    \multirow{2}*{SR-RAG} & 15.5 & 16.9  & 8.3 & 5.5 & \multirow{2}{*}{1.6}  & \multirow{2}{*}{16.08} \\
     & \% & $\textbf{9.0}^{\uparrow}$ & $50.9^{\downarrow}$ & $33.7^{\downarrow}$  \\
    \midrule
    \multirow{2}*{SEAKR} & 30.2 & 30.5 & 25.7 & 21.4 & \multirow{2}*{1.25} &  \multirow{2}{*}{24.03}\\
     & \%  & $\textbf{1.0}^{\uparrow}$ & $15.7^{\downarrow}$ & $16.7^{\downarrow}$ \\
    \midrule
    \multirow{2}*{\shortstack{\textbf{DSSP-}\\\textbf{RAG}}}
     & 28.4 & 31.6 & 32.1 & 31.3 & \multirow{2}*{0.72}  & \multirow{2}{*}{17.34} \\
     & \% & $\textbf{11.3}^{\uparrow}$ & $\textbf{1.6}^{\uparrow}$ & $2.5^{\downarrow}$ \\
    \bottomrule
    \end{tabular}
    \caption{Analysis of inference efficiency and document count of knowledge-enhanced methods.  Inf. (s) measures the time for autoregressive decoding only; Mem. (GB) indicates peak GPU memory usage.}
    \label{tab:knowNumber}
\vspace{-8pt}
\end{table}

As illustrated in Figure~\ref{hyperFigure}, we validate the effectiveness of DSSP-RAG's dynamic plug-in strategy, which selects the DSSP insertion layer based on the maximal distance between sampling subspaces of semantically equivalent but syntactically different prompts. Compared with fixed plug-in positions at early, middle, and late layers, the dynamic strategy consistently outperforms all fixed settings, including the best static configuration at the 25th layer, highlighting the layer-wise variability in knowledge encoding. Dynamically selecting the most activated layer enables more effective interaction with external knowledge and improves information extraction.
Moreover, we observe that when the number of similar tokens $T$ between internal and external knowledge is 10, the collaborative effect of shared and private semantics reaches its peak. Through grid search, we determine that DSSP-RAG achieves optimal performance with regularization coefficients of 0.55 for conditional entropy and 0.1 for KL divergence.

\begin{figure}[t!]
    \centering
    \subfloat{%
        \begin{tikzpicture}[font=\Large, scale=0.4]
        \begin{axis}[
            axis lines=box,
            axis x line=top,
            axis y line=left,
            every outer x axis line/.append style={blue},
            every outer y axis line/.append style={black},
            xlabel={Similar Token $T$},
            xtick pos=top,
            tick label style={font=\large},
            ylabel style={font=\large},
            ylabel={Exact Match},
            ytick style={font=\large},
            yticklabel style={font=\large},
            ymajorgrids=true,
            grid style=dashed,
            symbolic x coords={5,10,15,20},
            xtick=data,
            ymin=0, ymax=35,
            enlargelimits=true,
            legend pos=north west,
        ]
        \addplot[
            color=blue,
            mark=square*,
            line width=2pt,
            mark size=2pt
        ] coordinates {
            (5,22.3) (10,31.6) (15,23.7) (20,26.1)
        };
        \end{axis}
        \begin{axis}[
            xlabel={Layer $l$},
            xtick pos=bottom,
            axis x line=bottom,
            axis y line=right,
            ytick=\empty,
            yticklabels=\empty,
            every outer x axis line/.append style={red},
            tick label style={font=\large},
            symbolic x coords={5,15,25,28},
            xtick=data,
            ymin=0, ymax=35,
            enlargelimits=true,
        ]
        \addplot[
            color=red,
            mark=triangle*,
            line width=2pt,
            mark size=2pt
        ] coordinates {
            (5,3.5) (15,17.3) (25,28.7) (28,24.4)
        };
        \addplot[
            color=green,
            dashed,
            line width=2pt
        ] coordinates {
            (5,31.6) (28,31.6)
        };
        \end{axis}
        \end{tikzpicture}
    }
    \hfill
    \subfloat{%
        \begin{tikzpicture}[font=\Large, scale=0.4]
        \begin{axis}[
            axis lines=box,
            axis x line=top,
            axis y line=left,
            every outer x axis line/.append style={c5},
            every outer y axis line/.append style={black},
            xlabel={Regularization coefficient $\mu$ of $H(\cdot )$},
            xtick pos=top,
            tick label style={font=\large},
            ylabel style={font=\large},
            ylabel={Exact Match},
            ytick style={font=\large},
            yticklabel style={font=\large},
            ymajorgrids=true,
            grid style=dashed,
            symbolic x coords={0.1,0.4,0.5,0.52,0.55,0.57,0.6,0.7},
            xtick=data,
            ymin=18, ymax=32,
            enlargelimits=true,
        ]
        \addplot[
            color=c5,
            mark=square*,
            line width=2pt,
            mark size=2pt
        ] coordinates {
            (0.4,18.3) (0.5,27.9) (0.52,23.7) (0.55,31.6) (0.57,20.8) (0.6,28.4) (0.7,24.6)
        };
        \addlegendentry{$\nu=0.1$}
        \end{axis}
        \begin{axis}[
            xlabel={Regularization coefficient $\nu$ of $D_{KL}$},
            xtick pos=bottom,
            axis x line=bottom,
            axis y line=right,
            ytick=\empty,
            yticklabels=\empty,
            every outer x axis line/.append style={c1},
            tick label style={font=\large},
            symbolic x coords={0.08,0.1,0.11,0.12,0.13,0.15,0.2},
            xtick=data,
            ymin=18, ymax=32,
            enlargelimits=true,
            legend pos=south east,
        ]
        \addplot[
            color=c1,
            mark=triangle*,
            line width=2pt,
            mark size=2pt
        ] coordinates {
            (0.08,25.7) (0.1,31.6) (0.11,22.8) (0.12,24.5) (0.13,26.7) (0.15,28.3) (0.2,25.1)
        };
        \addlegendentry{$\mu=0.55$}
        \end{axis}
        \end{tikzpicture}
    }

    \caption{The analysis of hyperparameters.}
    \label{hyperFigure}
    \vspace{-16pt}
\end{figure}

\section{Conclusions}
\label{conclusion}

We have proposed DSSP-RAG to address the inter-context and context-memory knowledge conflicts in RAG. A mixed attention mechanism is developed to disentangle the internal and external knowledge into shared and private semantics, enabling fine-grained control over knowledge utilization.
An unsupervised hallucination detection based on sampling subspace divergence guides the integration of external knowledge, and an Energy Quotient is introduced to filter the noise via attention difference matrices. Extensive experiments demonstrate that DSSP-RAG significantly outperforms a wide range of the state-of-the-art baselines.

\section{Limitations}
\label{sec:limitation}
To mitigate the hallucination issues, DSSP-RAG incorporates hallucination detection, knowledge filtering, and a mixed-attention module for shared-private semantic synergy, enhancing the effectiveness of external knowledge augmentation in LLMs. Our hallucination detection method operates without annotated data or supervised training, identifying cognitive uncertainty in LLMs by assessing the stability of sampling subspaces constructed from internal states. This enables a precise determination of when external knowledge should be introduced. However, LLMs may occasionally generate consistent responses to syntactically different but semantically identical prompts due to coincidence, or they may produce stable but incorrect answers that deviate from the intended response. To address these limitations, we will propose further advancements in hallucination detection methods based on internal model states and a more in-depth investigation of LLMs' internal representations.  

Additionally, we evaluated the robustness of DSSP-RAG and the baseline models under varying numbers of external documents. Although DSSP-RAG shows a general performance improvement with an increasing number of retrieved documents, a slight decline was observed when the document count reached 20. This suggests potential interference from redundant or noisy information, leading to diminishing returns. In future research, we aim to refine external knowledge selection mechanisms by integrating more advanced noise filtering strategies and context-aware integration techniques to enhance the stability and generalization capabilities of DSSP-RAG under larger-scale external knowledge conditions.

\section*{Acknowledgments}
This work is funded in part by Natural Science Foundation of China (grant number: 62376027).


\bibliography{custom}

@article{huang2023survey,
  title={A survey on hallucination in large language models: Principles, taxonomy, challenges, and open questions},
  author={Huang, Lei and Yu, Weijiang and Ma, Weitao and Zhong, Weihong and Feng, Zhangyin and Wang, Haotian and Chen, Qianglong and Peng, Weihua and Feng, Xiaocheng and Qin, Bing and others},
  journal={arXiv preprint arXiv:2311.05232},
  year={2023}
}

@article{ram2023context,
  title={In-context retrieval-augmented language models},
  author={Ram, Ori and Levine, Yoav and Dalmedigos, Itay and Muhlgay, Dor and Shashua, Amnon and Leyton-Brown, Kevin and Shoham, Yoav},
  journal={Transactions of the Association for Computational Linguistics},
  volume={11},
  pages={1316--1331},
  year={2023},
  publisher={MIT Press One Broadway, 12th Floor, Cambridge, Massachusetts 02142, USA~…}
}

@article{shao2023enhancing,
  title={Enhancing retrieval-augmented large language models with iterative retrieval-generation synergy},
  author={Shao, Zhihong and Gong, Yeyun and Shen, Yelong and Huang, Minlie and Duan, Nan and Chen, Weizhu},
  journal={arXiv preprint arXiv:2305.15294},
  year={2023}
}

@article{jeong2024adaptive,
  title={Adaptive-rag: Learning to adapt retrieval-augmented large language models through question complexity},
  author={Jeong, Soyeong and Baek, Jinheon and Cho, Sukmin and Hwang, Sung Ju and Park, Jong C},
  journal={arXiv preprint arXiv:2403.14403},
  year={2024}
}

@article{he2024can,
  title={Can Language Models Act as Knowledge Bases at Scale?},
  author={He, Qiyuan and Wang, Yizhong and Wang, Wenya},
  journal={arXiv preprint arXiv:2402.14273},
  year={2024}
}

@article{wang2023survey,
  title={Survey on factuality in large language models: Knowledge, retrieval and domain-specificity},
  author={Wang, Cunxiang and Liu, Xiaoze and Yue, Yuanhao and Tang, Xiangru and Zhang, Tianhang and Jiayang, Cheng and Yao, Yunzhi and Gao, Wenyang and Hu, Xuming and Qi, Zehan and others},
  journal={arXiv preprint arXiv:2310.07521},
  year={2023}
}

@inproceedings{azaria2023internal,
  title={The internal state of an LLM knows when it's lying},
  author={Azaria, Amos and Mitchell, Tom},
  booktitle={Findings of the Association for Computational Linguistics: EMNLP 2023, Findings ’23},
  volume={1},
  pages={967–976},
  year={2023},
}

@article{kadavath2022language,
  title={Language models (mostly) know what they know},
  author={Kadavath, Saurav and Conerly, Tom and Askell, Amanda and Henighan, Tom and Drain, Dawn and Perez, Ethan and Schiefer, Nicholas and Hatfield-Dodds, Zac and DasSarma, Nova and Tran-Johnson, Eli and others},
  journal={arXiv preprint arXiv:2207.05221},
  year={2022}
}

@inproceedings{burns2022discovering,
  title={Discovering latent knowledge in language models without supervision},
  author={Burns, Collin and Ye, Haotian and Klein, Dan and Steinhardt, Jacob},
  booktitle={The Eleventh International Conference on Learning Representations, ICLR ’23},
  year={2023},
}

@inproceedings{xiong2023can,
  title={Can llms express their uncertainty? an empirical evaluation of confidence elicitation in llms},
  author={Xiong, Miao and Hu, Zhiyuan and Lu, Xinyang and Li, Yifei and Fu, Jie and He, Junxian and Hooi, Bryan},
  booktitle={The Twelfth International Conference on Learning Representations, ICLR ’24},
  year={2024},
}

@inproceedings{yin2023large,
  title={Do Large Language Models Know What They Don't Know?},
  author={Yin, Zhangyue and Sun, Qiushi and Guo, Qipeng and Wu, Jiawen and Qiu, Xipeng and Huang, Xuanjing},
  booktitle={Findings of the Association for Computational Linguistics: ACL 2023, Findings ’23},
  year={2023},
}

@article{lin2022teaching,
  title={Teaching models to express their uncertainty in words},
  author={Lin, Stephanie and Hilton, Jacob and Evans, Owain},
  journal={arXiv preprint arXiv:2205.14334},
  year={2022}
}

@inproceedings{tian2023just,
  title={Just ask for calibration: Strategies for eliciting calibrated confidence scores from language models fine-tuned with human feedback},
  author={Tian, Katherine and Mitchell, Eric and Zhou, Allan and Sharma, Archit and Rafailov, Rafael and Yao, Huaxiu and Finn, Chelsea and Manning, Christopher D},
  booktitle={Proceedings of the 2023 Conference on Empirical Methods in Natural Language Processing, EMNLP ’23},
  pages={967–976},
  year={2023},
}

@inproceedings{wang2022self,
  title={Self-consistency improves chain of thought reasoning in language models},
  author={Wang, Xuezhi and Wei, Jason and Schuurmans, Dale and Le, Quoc and Chi, Ed and Narang, Sharan and Chowdhery, Aakanksha and Zhou, Denny},
  booktitle={The Eleventh International Conference on Learning Representations, ICLR ’23},
  year={2023},
}

@inproceedings{manakul2023selfcheckgpt,
  title={Selfcheckgpt: Zero-resource black-box hallucination detection for generative large language models},
  author={Manakul, Potsawee and Liusie, Adian and Gales, Mark JF},
  booktitle={Proceedings of the 2023 Conference on Empirical Methods in Natural Language Processing, EMNLP ’23},
  pages={9004–9017},
  year={2023},
}

@inproceedings{kuhn2023semantic,
  title={Semantic uncertainty: Linguistic invariances for uncertainty estimation in natural language generation},
  author={Kuhn, Lorenz and Gal, Yarin and Farquhar, Sebastian},
  booktitle={The Eleventh International Conference on Learning Representations, ICLR ’23},
  year={2023},
}

@article{mahaut2024factual,
  title={Factual Confidence of LLMs: on Reliability and Robustness of Current Estimators},
  author={Mahaut, Mat{\'e}o and Aina, Laura and Czarnowska, Paula and Hardalov, Momchil and M{\"u}ller, Thomas and M{\`a}rquez, Llu{\'\i}s},
  journal={arXiv preprint arXiv:2406.13415},
  year={2024}
}

@article{yang2018hotpotqa,
  title={HotpotQA: A dataset for diverse, explainable multi-hop question answering},
  author={Yang, Zhilin and Qi, Peng and Zhang, Saizheng and Bengio, Yoshua and Cohen, William W and Salakhutdinov, Ruslan and Manning, Christopher D},
  journal={arXiv preprint arXiv:1809.09600},
  year={2018}
}

@article{ho2020constructing,
  title={Constructing a multi-hop qa dataset for comprehensive evaluation of reasoning steps},
  author={Ho, Xanh and Nguyen, Anh-Khoa Duong and Sugawara, Saku and Aizawa, Akiko},
  journal={arXiv preprint arXiv:2011.01060},
  year={2020}
}

@article{kwiatkowski2019natural,
  title={Natural questions: a benchmark for question answering research},
  author={Kwiatkowski, Tom and Palomaki, Jennimaria and Redfield, Olivia and Collins, Michael and Parikh, Ankur and Alberti, Chris and Epstein, Danielle and Polosukhin, Illia and Devlin, Jacob and Lee, Kenton and others},
  journal={Transactions of the Association for Computational Linguistics},
  volume={7},
  pages={453--466},
  year={2019},
  publisher={MIT Press One Rogers Street, Cambridge, MA 02142-1209, USA journals-info~…}
}

@inproceedings{joshi2017triviaqa,
  title={Triviaqa: A large scale distantly supervised challenge dataset for reading comprehension},
  author={Joshi, Mandar and Choi, Eunsol and Weld, Daniel S and Zettlemoyer, Luke},
  booktitle={Proceedings of the 55th Annual Meeting of the Association for Computational Linguistics, ACL 2017},
  volume={1},
  pages={1601–1611},
  year={2017},
}

@article{touvron2023llama,
  title={Llama 2: Open foundation and fine-tuned chat models},
  author={Touvron, Hugo and Martin, Louis and Stone, Kevin and Albert, Peter and Almahairi, Amjad and Babaei, Yasmine and Bashlykov, Nikolay and Batra, Soumya and Bhargava, Prajjwal and Bhosale, Shruti and others},
  journal={arXiv preprint arXiv:2307.09288},
  year={2023}
}

@article{asai2023self,
  title={Self-rag: Learning to retrieve, generate, and critique through self-reflection},
  author={Asai, Akari and Wu, Zeqiu and Wang, Yizhong and Sil, Avirup and Hajishirzi, Hannaneh},
  journal={arXiv preprint arXiv:2310.11511},
  year={2023}
}

@article{jiang2023active,
  title={Active retrieval augmented generation},
  author={Jiang, Zhengbao and Xu, Frank F and Gao, Luyu and Sun, Zhiqing and Liu, Qian and Dwivedi-Yu, Jane and Yang, Yiming and Callan, Jamie and Neubig, Graham},
  journal={arXiv preprint arXiv:2305.06983},
  year={2023}
}

@article{su2024dragin,
  title={Dragin: Dynamic retrieval augmented generation based on the real-time information needs of large language models},
  author={Su, Weihang and Tang, Yichen and Ai, Qingyao and Wu, Zhijing and Liu, Yiqun},
  journal={arXiv preprint arXiv:2403.10081},
  year={2024}
}

@article{yao2024seakr,
  title={Seakr: Self-aware knowledge retrieval for adaptive retrieval augmented generation},
  author={Yao, Zijun and Qi, Weijian and Pan, Liangming and Cao, Shulin and Hu, Linmei and Liu, Weichuan and Hou, Lei and Li, Juanzi},
  journal={arXiv preprint arXiv:2406.19215},
  year={2024}
}

@inproceedings{trivedi2022interleaving,
  title={Interleaving retrieval with chain-of-thought reasoning for knowledge-intensive multi-step questions},
  author={Trivedi, Harsh and Balasubramanian, Niranjan and Khot, Tushar and Sabharwal, Ashish},
  booktitle={In Proceedings of the 61st Annual Meeting of the Association for Computational Linguistics},
  volume={1},
  year={2022},
}

@article{robertson2009probabilistic,
  title={The probabilistic relevance framework: BM25 and beyond},
  author={Robertson, Stephen and Zaragoza, Hugo and others},
  journal={Foundations and Trends{\textregistered} in Information Retrieval},
  volume={3},
  number={4},
  pages={333--389},
  year={2009},
  publisher={Now Publishers, Inc.}
}

@article{meta2024introducing,
  title={Introducing meta llama 3: The most capable openly available llm to date, 2024},
  author={Meta, AI},
  journal={URL https://ai. meta. com/blog/meta-llama-3/. Accessed on April},
  volume={26},
  year={2024}
}

@inproceedings{wu2024clasheval,
  title={Clasheval: Quantifying the tug-of-war between an llm’s internal prior and external evidence},
  author={Wu, Kevin and Wu, Eric and Zou, James},
  booktitle={The Thirty-eight Conference on Neural Information Processing Systems Datasets and Benchmarks Track},
  year={2024}
}

@article{xu2024knowledge,
  title={Knowledge conflicts for llms: A survey},
  author={Xu, Rongwu and Qi, Zehan and Guo, Zhijiang and Wang, Cunxiang and Wang, Hongru and Zhang, Yue and Xu, Wei},
  journal={arXiv preprint arXiv:2403.08319},  
  year={2024}
}

@article{ni2024llms,
  title={When Do LLMs Need Retrieval Augmentation? Mitigating LLMs' Overconfidence Helps Retrieval Augmentation},
  author={Ni, Shiyu and Bi, Keping and Guo, Jiafeng and Cheng, Xueqi},
  journal={arXiv preprint arXiv:2402.11457},
  year={2024}
}

@article{meng2022mass,
  title={Mass-editing memory in a transformer},
  author={Meng, Kevin and Sharma, Arnab Sen and Andonian, Alex and Belinkov, Yonatan and Bau, David},
  journal={arXiv preprint arXiv:2210.07229},
  year={2022}
}

@article{fan2025research,
  title={Research on the Online Update Method for Retrieval-Augmented Generation (RAG) Model with Incremental Learning},
  author={Fan, Yuxin and Wang, Yuxiang and Liu, Lipeng and Tang, Xirui and Sun, Na and Yu, Zidong},
  journal={arXiv preprint arXiv:2501.07063},
  year={2025}
}

@article{tonmoy2024comprehensive,
  title={A comprehensive survey of hallucination mitigation techniques in large language models},
  author={Tonmoy, SM and Zaman, SM and Jain, Vinija and Rani, Anku and Rawte, Vipula and Chadha, Aman and Das, Amitava},
  journal={arXiv preprint arXiv:2401.01313},
  year={2024}
}

@article{huang2024survey,
  title={A survey on hallucination in large language models: Principles, taxonomy, challenges, and open questions},
  author={Huang, Lei and Yu, Weijiang and Ma, Weitao and Zhong, Weihong and Feng, Zhangyin and Wang, Haotian and Chen, Qianglong and Peng, Weihua and Feng, Xiaocheng and Qin, Bing and others},
  journal={ACM Transactions on Information Systems},
  year={2024},
  publisher={ACM New York, NY}
}

@article{qian2024memorag,
  title={Memorag: Moving towards next-gen rag via memory-inspired knowledge discovery},
  author={Qian, Hongjin and Zhang, Peitian and Liu, Zheng and Mao, Kelong and Dou, Zhicheng},
  journal={arXiv preprint arXiv:2409.05591},
  year={2024}
}

@article{yu2025rankrag,
  title={Rankrag: Unifying context ranking with retrieval-augmented generation in llms},
  author={Yu, Yue and Ping, Wei and Liu, Zihan and Wang, Boxin and You, Jiaxuan and Zhang, Chao and Shoeybi, Mohammad and Catanzaro, Bryan},
  journal={Advances in Neural Information Processing Systems},
  volume={37},
  pages={121156--121184},
  year={2025}
}

@article{yadkori2024believe,
  title={To Believe or Not to Believe Your LLM},
  author={Yadkori, Yasin Abbasi and Kuzborskij, Ilja and Gy{\"o}rgy, Andr{\'a}s and Szepesv{\'a}ri, Csaba},
  journal={arXiv preprint arXiv:2406.02543},
  year={2024}
}

@article{wang2024blendfilter,
  title={Blendfilter: Advancing retrieval-augmented large language models via query generation blending and knowledge filtering},
  author={Wang, Haoyu and Li, Ruirui and Jiang, Haoming and Tian, Jinjin and Wang, Zhengyang and Luo, Chen and Tang, Xianfeng and Cheng, Monica and Zhao, Tuo and Gao, Jing},
  journal={arXiv preprint arXiv:2402.11129},
  year={2024}
}

@inproceedings{fan2024survey,
  title={A survey on rag meeting llms: Towards retrieval-augmented large language models},
  author={Fan, Wenqi and Ding, Yujuan and Ning, Liangbo and Wang, Shijie and Li, Hengyun and Yin, Dawei and Chua, Tat-Seng and Li, Qing},
  booktitle={Proceedings of the 30th ACM SIGKDD Conference on Knowledge Discovery and Data Mining},
  pages={6491--6501},
  year={2024}
}

@article{ji2024llm,
  title={Llm internal states reveal hallucination risk faced with a query},
  author={Ji, Ziwei and Chen, Delong and Ishii, Etsuko and Cahyawijaya, Samuel and Bang, Yejin and Wilie, Bryan and Fung, Pascale},
  journal={arXiv preprint arXiv:2407.03282},
  year={2024}
}

@article{zhou2024trustworthiness,
  title={Trustworthiness in retrieval-augmented generation systems: A survey},
  author={Zhou, Yujia and Liu, Yan and Li, Xiaoxi and Jin, Jiajie and Qian, Hongjin and Liu, Zheng and Li, Chaozhuo and Dou, Zhicheng and Ho, Tsung-Yi and Yu, Philip S},
  journal={arXiv preprint arXiv:2409.10102},
  year={2024}
}

@article{yu2023improving,
  title={Improving language models via plug-and-play retrieval feedback},
  author={Yu, Wenhao and Zhang, Zhihan and Liang, Zhenwen and Jiang, Meng and Sabharwal, Ashish},
  journal={arXiv preprint arXiv:2305.14002},
  year={2023}
}

@article{zhang2024comprehensive,
  title={A comprehensive study of knowledge editing for large language models},
  author={Zhang, Ningyu and Yao, Yunzhi and Tian, Bozhong and Wang, Peng and Deng, Shumin and Wang, Mengru and Xi, Zekun and Mao, Shengyu and Zhang, Jintian and Ni, Yuansheng and others},
  journal={arXiv preprint arXiv:2401.01286},
  year={2024}
}

@article{zhang2024truthx,
  title={Truthx: Alleviating hallucinations by editing large language models in truthful space},
  author={Zhang, Shaolei and Yu, Tian and Feng, Yang},
  journal={arXiv preprint arXiv:2402.17811},
  year={2024}
}

@article{hase2024does,
  title={Does localization inform editing? surprising differences in causality-based localization vs. knowledge editing in language models},
  author={Hase, Peter and Bansal, Mohit and Kim, Been and Ghandeharioun, Asma},
  journal={Advances in Neural Information Processing Systems},
  volume={36},
  year={2024}
}

@article{wang2024knowledge,
  title={Knowledge Editing through Chain-of-Thought},
  author={Wang, Changyue and Su, Weihang and Ai, Qingyao and Liu, Yiqun},
  journal={arXiv preprint arXiv:2412.17727},
  year={2024}
}

@inproceedings{du2022syntheticdisinformationattacksautomated,
      title={Synthetic Disinformation Attacks on Automated Fact Verification Systems}, 
      author={Yibing Du and Antoine Bosselut and Christopher D. Manning},
booktitle={In Proceedings of the AAAI Conference on Artificial Intelligence},
volume={36},
pages={10581–10589},
  year={2022}
}

@inproceedings{pan-etal-2023-attacking,
    title={Attacking Open-domain Question Answering by Injecting Misinformation},
    author={ Pan, Liangming  and  Chen, Wenhu  and  Kan, Min Yen  and  Wang, William Yang },
    booktitle = {Proceedings of the 13th International Joint Conference on Natural Language Processing and the 3rd Conference of the Asia-Pacific Chapter of the Association for Computational Linguistics (Volume 1: Long Papers)},
    month = nov,
    year = "2023",
    address = "Nusa Dua, Bali",
    publisher = "Association for Computational Linguistics",
    url = "https://aclanthology.org/2023.ijcnlp-main.35/",
    doi = "10.18653/v1/2023.ijcnlp-main.35",
    pages = "525--539"
}

@misc{jin2024longcontextllmsmeetrag,
      title={Long-Context LLMs Meet RAG: Overcoming Challenges for Long Inputs in RAG}, 
      author={Bowen Jin and Jinsung Yoon and Jiawei Han and Sercan O. Arik},
      year={2024},
      eprint={2410.05983},
      archivePrefix={arXiv},
      primaryClass={cs.CL},
      url={https://arxiv.org/abs/2410.05983}, 
}

@inproceedings{Dong2025Decouplingknowledgeandcontext,
author = {Dong, Qian and Ai, Qingyao and Wang, Hongning and Liu, Yiding and Li, Haitao and Su, Weihang and Liu, Yiqun and Chua, Tat-Seng and Ma, Shaoping},
title = {Decoupling Knowledge and Context: An Efficient and Effective Retrieval Augmented Generation Framework via Cross Attention},
year = {2025},
isbn = {9798400712746},
publisher = {Association for Computing Machinery},
address = {New York, NY, USA},
booktitle = {Proceedings of the ACM on Web Conference 2025},
pages = {4386–4395},
numpages = {10},
}

@inproceedings{yang2024gatedlinearattentiontransformers,
author = {Yang, Songlin and Wang, Bailin and Shen, Yikang and Panda, Rameswar and Kim, Yoon},
title = {Gated linear attention transformers with hardware-efficient training},
year = {2024},
publisher = {JMLR.org},
booktitle = {Proceedings of the 41st International Conference on Machine Learning},
articleno = {2333},
numpages = {23},
location = {Vienna, Austria},
series = {ICML'24}
}

@inproceedings{zhang2023crashclusteringremovingsharing,
      title={CRaSh: Clustering, Removing, and Sharing Enhance Fine-tuning without Full Large Language Model}, 
      author={Kaiyan Zhang and Ning Ding and Biqing Qi and Xuekai Zhu and Xinwei Long and Bowen Zhou},
    booktitle = {Proceedings of the 2023 Conference on Empirical Methods in Natural Language Processing},
    month = dec,
    year = {2023},
    address = {Singapore},
    publisher = {Association for Computational Linguistics},
    doi = {10.18653/v1/2023.emnlp-main.597},
    pages = {9612--9637},
    
}

@misc{ju2024largelanguagemodelsencode,
      title={How Large Language Models Encode Context Knowledge? A Layer-Wise Probing Study}, 
      author={Tianjie Ju and Weiwei Sun and Wei Du and Xinwei Yuan and Zhaochun Ren and Gongshen Liu},
      year={2024},
      eprint={2402.16061},
      archivePrefix={arXiv},
      primaryClass={cs.CL},
      url={https://arxiv.org/abs/2402.16061}, 
}

@misc{xie2024adaptivechameleonstubbornsloth,
      title={Adaptive Chameleon or Stubborn Sloth: Revealing the Behavior of Large Language Models in Knowledge Conflicts}, 
      author={Jian Xie and Kai Zhang and Jiangjie Chen and Renze Lou and Yu Su},
      year={2024},
      eprint={2305.13300},
      archivePrefix={arXiv},
      primaryClass={cs.CL},
      url={https://arxiv.org/abs/2305.13300}, 
}

@article{liu-etal-2024-lost,
    title = "Lost in the Middle: How Language Models Use Long Contexts",
    author = "Liu, Nelson F.  and
      Lin, Kevin  and
      Hewitt, John  and
      Paranjape, Ashwin  and
      Bevilacqua, Michele  and
      Petroni, Fabio  and
      Liang, Percy",
    journal = "Transactions of the Association for Computational Linguistics",
    volume = "12",
    year = "2024",
    address = "Cambridge, MA",
    publisher = "MIT Press",
    doi = "10.1162/tacl_a_00638",
    pages = "157--173"
}

@misc{kotonya2020explainableautomatedfactcheckingpublic,
      title={Explainable Automated Fact-Checking for Public Health Claims}, 
      author={Neema Kotonya and Francesca Toni},
      year={2020},
      eprint={2010.09926},
      archivePrefix={arXiv},
      primaryClass={cs.CL},
      url={https://arxiv.org/abs/2010.09926}, 
}

@misc{qwen2025qwen25technicalreport,
      title={Qwen2.5 Technical Report}, 
      author={Qwen and : and An Yang and Baosong Yang and Beichen Zhang and Binyuan Hui and Bo Zheng and Bowen Yu and Chengyuan Li and Dayiheng Liu and Fei Huang and Haoran Wei and Huan Lin and Jian Yang and Jianhong Tu and Jianwei Zhang and Jianxin Yang and Jiaxi Yang and Jingren Zhou and Junyang Lin and Kai Dang and Keming Lu and Keqin Bao and Kexin Yang and Le Yu and Mei Li and Mingfeng Xue and Pei Zhang and Qin Zhu and Rui Men and Runji Lin and Tianhao Li and Tianyi Tang and Tingyu Xia and Xingzhang Ren and Xuancheng Ren and Yang Fan and Yang Su and Yichang Zhang and Yu Wan and Yuqiong Liu and Zeyu Cui and Zhenru Zhang and Zihan Qiu},
      year={2025},
      eprint={2412.15115},
      archivePrefix={arXiv},
      primaryClass={cs.CL},
      url={https://arxiv.org/abs/2412.15115}, 
}

@misc{wang2025chainofretrievalaugmentedgeneration,
      title={Chain-of-Retrieval Augmented Generation}, 
      author={Liang Wang and Haonan Chen and Nan Yang and Xiaolong Huang and Zhicheng Dou and Furu Wei},
      year={2025},
      eprint={2501.14342},
      archivePrefix={arXiv},
      primaryClass={cs.IR},
      url={https://arxiv.org/abs/2501.14342}, 
}

@misc{lee2025rearagknowledgeguidedreasoningenhances,
      title={ReaRAG: Knowledge-guided Reasoning Enhances Factuality of Large Reasoning Models with Iterative Retrieval Augmented Generation}, 
      author={Zhicheng Lee and Shulin Cao and Jinxin Liu and Jiajie Zhang and Weichuan Liu and Xiaoyin Che and Lei Hou and Juanzi Li},
      year={2025},
      eprint={2503.21729},
      archivePrefix={arXiv},
      primaryClass={cs.CL},
      url={https://arxiv.org/abs/2503.21729}, 
}

@misc{gao2025synergizingragreasoningsystematic,
      title={Synergizing RAG and Reasoning: A Systematic Review}, 
      author={Yunfan Gao and Yun Xiong and Yijie Zhong and Yuxi Bi and Ming Xue and Haofen Wang},
      year={2025},
      eprint={2504.15909},
      archivePrefix={arXiv},
      primaryClass={cs.IR},
      url={https://arxiv.org/abs/2504.15909}, 
}

@misc{guan2025deepragthinkingretrievalstep,
      title={DeepRAG: Thinking to Retrieval Step by Step for Large Language Models}, 
      author={Xinyan Guan and Jiali Zeng and Fandong Meng and Chunlei Xin and Yaojie Lu and Hongyu Lin and Xianpei Han and Le Sun and Jie Zhou},
      year={2025},
      eprint={2502.01142},
      archivePrefix={arXiv},
      primaryClass={cs.AI},
      url={https://arxiv.org/abs/2502.01142}, 
}

@article{ding2024retrieve,
  title={Retrieve only when it needs: Adaptive retrieval augmentation for hallucination mitigation in large language models},
  author={Ding, Hanxing and Pang, Liang and Wei, Zihao and Shen, Huawei and Cheng, Xueqi},
  journal={arXiv preprint arXiv:2402.10612},
  year={2024}
}

@article{peng2025cafe,
  title={CAFE: Retrieval Head-based Coarse-to-Fine Information Seeking to Enhance Multi-Document QA Capability},
  author={Peng, Han and Jiang, Jinhao and Dong, Zican and Zhao, Wayne Xin and Fang, Lei},
  journal={arXiv preprint arXiv:2505.10063},
  year={2025}
}

@inproceedings{qiu2024entropy,
  title={Entropy-based decoding for retrieval-augmented large language models},
  author={Qiu, Zexuan and Ou, Zijing and Wu, Bin and Li, Jingjing and Liu, Aiwei and King, Irwin},
  booktitle={Proceedings of the 2025 Conference of the Nations of the Americas Chapter of the Association for Computational Linguistics: Human Language Technologies (Volume 1: Long Papers)},
  pages={4616--4627},
  year={2025}
}

@article{bi2025parameters,
  title={Parameters vs. context: Fine-grained control of knowledge reliance in language models},
  author={Bi, Baolong and Liu, Shenghua and Wang, Yiwei and Xu, Yilong and Fang, Junfeng and Mei, Lingrui and Cheng, Xueqi},
  journal={arXiv preprint arXiv:2503.15888},
  year={2025}
}

@article{li2025loki,
  title={Loki's Dance of Illusions: A Comprehensive Survey of Hallucination in Large Language Models},
  author={Li, Chaozhuo and Wang, Pengbo and Wang, Chenxu and Zhang, Litian and Liu, Zheng and Ye, Qiwei and Xu, Yuanbo and Huang, Feiran and Zhang, Xi and Yu, Philip S},
  journal={arXiv preprint arXiv:2507.02870},
  year={2025}
}

@article{wang2025two,
  title={Two Birds with One Stone: Improving Factuality and Faithfulness of LLMs via Dynamic Interactive Subspace Editing},
  author={Wang, Pengbo and Li, Chaozhuo and Wang, Chenxu and Zheng, Liwen and Zhang, Litian and Zhang, Xi},
  journal={arXiv preprint arXiv:2506.11088},
  year={2025}
}

@article{li2024generalizing,
  title={Generalizing knowledge graph embedding with universal orthogonal parameterization},
  author={Li, Rui and Li, Chaozhuo and Shen, Yanming and Zhang, Zeyu and Chen, Xu},
  journal={arXiv preprint arXiv:2405.08540},
  year={2024}
}

@inproceedings{li2023copy,
  title={To copy rather than memorize: A vertical learning paradigm for knowledge graph completion},
  author={Li, Rui and Chen, Xu and Li, Chaozhuo and Shen, Yanming and Zhao, Jianan and Wang, Yujing and Han, Weihao and Sun, Hao and Deng, Weiwei and Zhang, Qi and others},
  booktitle={Proceedings of the 61st Annual Meeting of the Association for Computational Linguistics (Volume 1: Long Papers)},
  pages={6335--6347},
  year={2023}
}

@inproceedings{li2022house,
  title={House: Knowledge graph embedding with householder parameterization},
  author={Li, Rui and Zhao, Jianan and Li, Chaozhuo and He, Di and Wang, Yiqi and Liu, Yuming and Sun, Hao and Wang, Senzhang and Deng, Weiwei and Shen, Yanming and others},
  booktitle={International conference on machine learning},
  pages={13209--13224},
  year={2022},
  organization={PMLR}
}

@inproceedings{stelmakh2022asqa,
  title={ASQA: Factoid questions meet long-form answers},
  author={Stelmakh, Ivan and Luan, Yi and Dhingra, Bhuwan and Chang, Ming-Wei},
  booktitle={Proceedings of the 2022 Conference on Empirical Methods in Natural Language Processing},
  pages={8273--8288},
  year={2022}
}

\appendix

\begin{table*}[h!]
\centering
\begin{tabular}{l|cc|cc|c}
\toprule

& \multicolumn{2}{c|}{\textbf{milti-hop QA}} & \multicolumn{2}{c}{\textbf{single-hop QA}} & closed-set of public health \\ 
\cline{2-3} \cline{4-5} 
\noalign{\vskip 1mm}
& \textbf{HotpotQA} & \textbf{2WikiMultihopQA} & \textbf{NQ} & \textbf{TriviaQA} & \textbf{PubHealth}  \\ 
\midrule
\#Examples & 1000 & 1000 & 3,610 & 11,313 & 11,800 \\
\bottomrule
\end{tabular}
\caption{The statistics of inference dataset.}
\label{tab:dataDetail}
\end{table*}

\section{Datasets and Settings}
\label{sec:dataAppend}
The fine-tuning dataset for DSSP-RAG consists of 30,000 samples from the train set of HotpotQA and 10,000 samples from the train set of 2WikiMultihopQA. For hyperparameter tuning, we utilize a subset of 3,000 HotpotQA samples and 2,000 2WikiMultihopQA samples. The details of the inference dataset statistics are shown in Table~\ref{tab:dataDetail}. And, the hyperparameters of DSSP-RAG on are listed in Table ~\ref{tab:hyperDetail}.

The tuning process of the regularization coefficients follows a systematic grid search strategy to determine the optimal parameter combination:
\begin{itemize}
    \item The KL divergence regularization term $\nu$ is introduced to constrain distributional shifts when integrating external knowledge, preventing the model from becoming overly reliant on external sources, which could lead to prediction biases. Given that a smaller initial value is typically preferable to avoid excessive constraints, $\nu$ is initialized at 0.12. The search space for $\nu$ is set within the range $[0.05,0.15]$ with a step size of 0.01.
    \item The conditional entropy regularization term $\mu$ is designed to mitigate uncertainty when leveraging external knowledge, thereby improving knowledge utilization efficiency. A relatively larger initial value helps the model reduce uncertainty more effectively, facilitating improved accuracy in knowledge integration. Accordingly, $\mu$ is initialized at 0.5, with a coarse search space defined as $[0.4,0.7]$ using a step size of 0.1. For finer adjustments, the step size is refined to 0.01 within the range $\mu \in [0.5,0.6]$.
\end{itemize}

To ensure optimal parameter selection, cross-validation is performed in the neighborhood of the most promising parameter $(0.1\pm 0.02,0.55\pm0.03)$. The results confirm that the globally optimal combination is $\nu=0.1$ and $\mu=0.55$.

\begin{table}[h!]
\centering
\begin{tabular}{l c} 
\toprule
\textbf{Hyperparameters} & \textbf{Value} \\ 
\midrule
JSD threshold $\delta$ & 1.0  \\
Similar token $T$ & 10  \\
Regularization coefficient $\mu$ & 0.55  \\
Regularization coefficient $\nu$ & 0.1  \\
Learning rate &  4e-5 \\
Epoch &  7 \\
Warm-up ratio &  0.1 \\
\bottomrule
\end{tabular}
\caption{Hyperparameters of DSSP-RAG.}
\label{tab:hyperDetail}
\end{table}

\section{Evaluation on the ASQA Dataset}
\label{sec:ASQAResult}

ASQA~\cite{stelmakh2022asqa} transforms QA into a summarization-style generation task, thereby allowing us to evaluate DSSP-RAG's capabilities in scenarios that demand both factual grounding and discourse-level generation quality. The results shown in Table~\ref{tab:ASQAResult} further demonstrate the extensibility of DSSP-RAG beyond traditional QA and reaffirm the effectiveness of our proposed modules under more generative settings.

\begin{table}[t]
\centering
\small
\begin{tabular}{l|cc|cc}
\toprule
\textbf{Model}      & \multicolumn{2}{c|}{\textbf{\makecell{LLaMA2\\-7B-Chat}}} & \multicolumn{2}{c}{\textbf{\makecell{LLaMA3\\-8B-Instruct}}} \\
\midrule
\textbf{Method}     & \textbf{str-em} & \textbf{R-L} & \textbf{str-em} & \textbf{R-L} \\
\midrule
\emph{Simple RAG}   &                 &              &                 &              \\
No Retrieval        & 16.7            & 15.5         & 29.7            & 33.4         \\
SR-RAG              & 24.2            & 29.6         & 35.0            & 36.3         \\
\textbf{DSSP\_SR}   & \textbf{25.7}   & \textbf{30.2}& \textbf{35.8}   & \textbf{37.9}\\
\midrule
\emph{Adaptive RAG} &                 &              &                 &              \\
Self-RAG            & 30.0            & 35.7         & 36.3            & 38.4         \\
FLARE               & 28.5            & 33.6         & 34.3            & 35.1         \\
SEAKR               & 30.8            & 36.4         & 40.1            & 40.6         \\
\textbf{DSSP\_AR}   & \textbf{32.2}   & \textbf{38.1}& \textbf{42.9}   & \textbf{43.3}\\
\midrule
\textbf{$\Delta$ (\%)} & $\uparrow$\textbf{4.54} & $\uparrow$\textbf{4.67} & $\uparrow$\textbf{6.98} & $\uparrow$\textbf{6.65} \\
\bottomrule
\end{tabular}
\caption{Performance comparison on ASQA dataset.}
\label{tab:ASQAResult}
\end{table}

\section{The details of hallucination detection}
\label{sec:halluAppend}

\begin{table*}[ht]
\centering
\begin{tabular}{p{0.35\textwidth}p{0.35\textwidth}p{0.25\textwidth}}
\toprule
\textbf{Prompt} & \textbf{Label} & \textbf{Model Answer} \\
\midrule
Where was the place of death of Yazid III's father? & \multirow{2}*{Damascus} & Damascus \\
father of Yazid III, where was place of death of him? &  & Damascus, Syria \\
\midrule
Where did the director of film Akcja Pod Arsenałem graduate from? & \multirow{2}*{National Film School in Łódź} & University of Warsaw \\
film Akcja Pod Arsenałem's director, where did him graduate from? & & Poland \\
\bottomrule
\end{tabular}
\caption{Comparison of Model Responses to Different Prompts for Two Questions}
\label{tab:PromptAnswer}
\end{table*}

\begin{figure*}
  \centering
    \begin{subfigure}[b]{0.48\textwidth}
        \includegraphics[width=\textwidth]{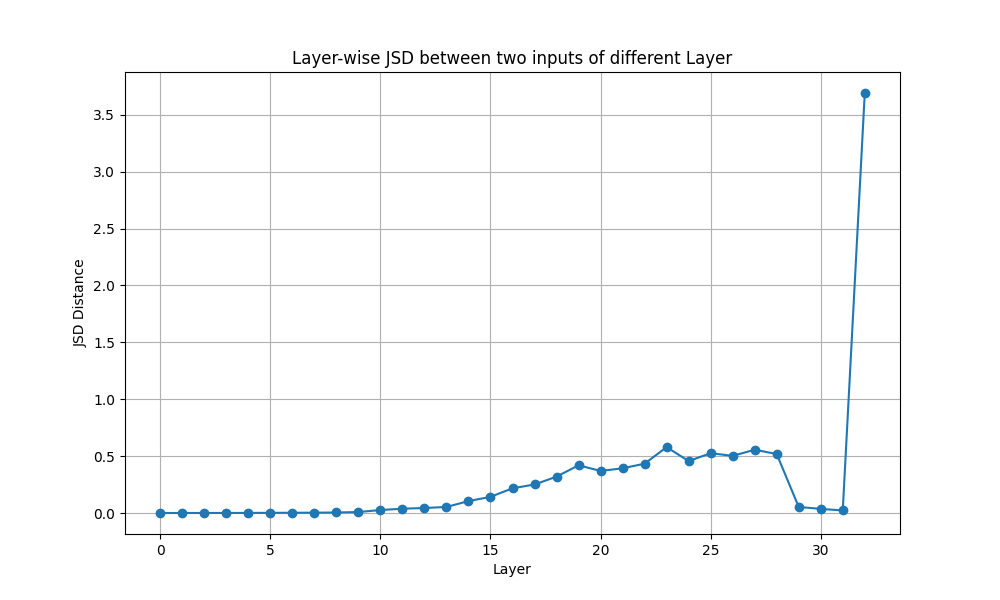} 
    \caption{Non-Hallucination}
    \label{fig:Nohallu}
    \end{subfigure}
    \hfill
    \begin{subfigure}[b]{0.48\textwidth}
        \includegraphics[width=\textwidth]{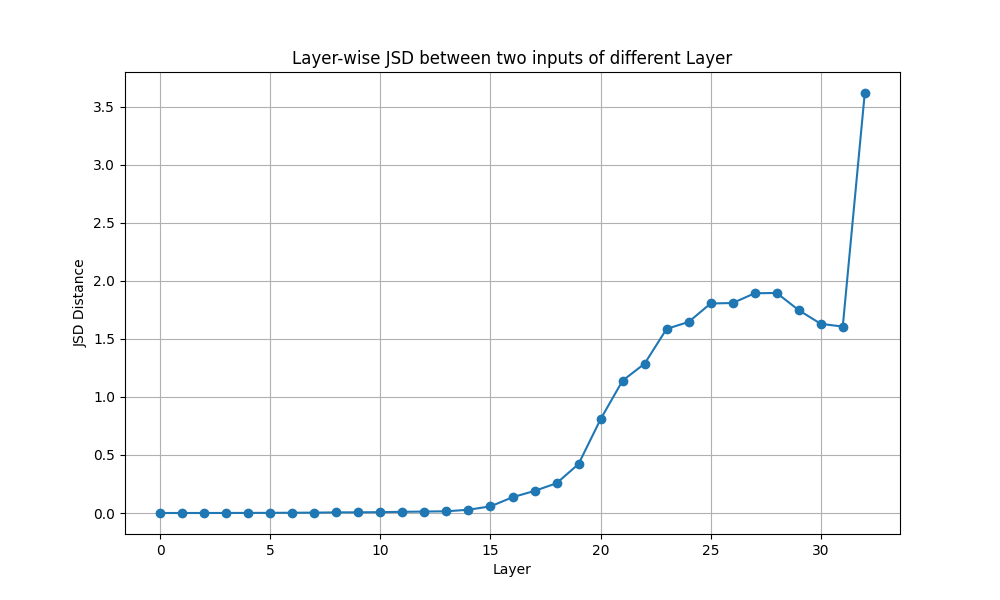}
    \caption{Hallucination}
    \label{fig:hallu}
    \end{subfigure}
    
    \caption{JSD Distances Between Sampling Subspaces of different prompts Under Non-Hallucination and Hallucination Scenarios in LLMs.}
    \label{fig:JSD}

\end{figure*}

\begin{figure*}
  \centering
    \begin{subfigure}[b]{0.48\textwidth}
        \includegraphics[width=\textwidth]{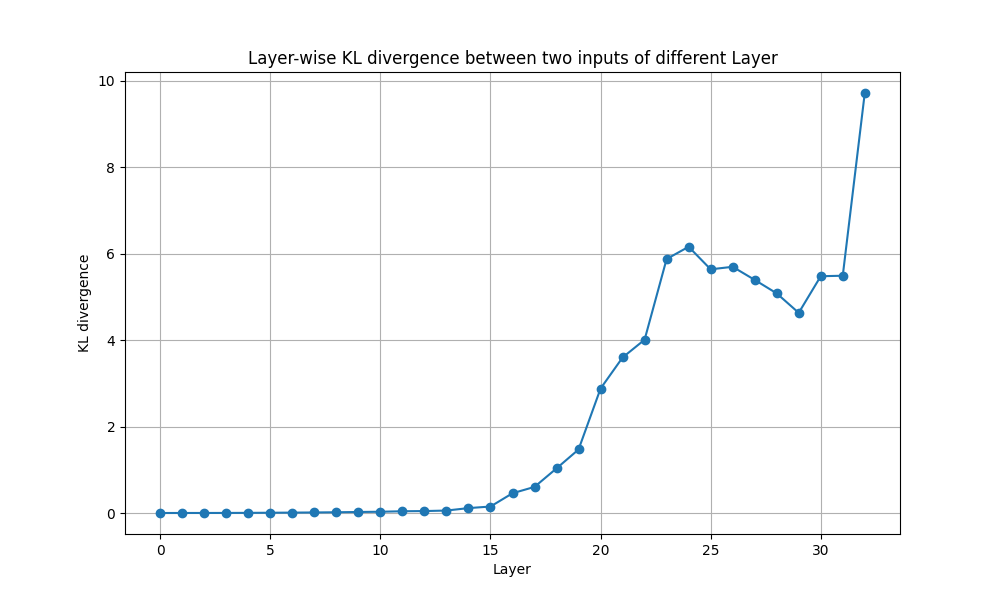} 
    \caption{Non-Hallucination}
    \label{fig:Nohallu}
    \end{subfigure}
    \hfill
    \begin{subfigure}[b]{0.48\textwidth}
        \includegraphics[width=\textwidth]{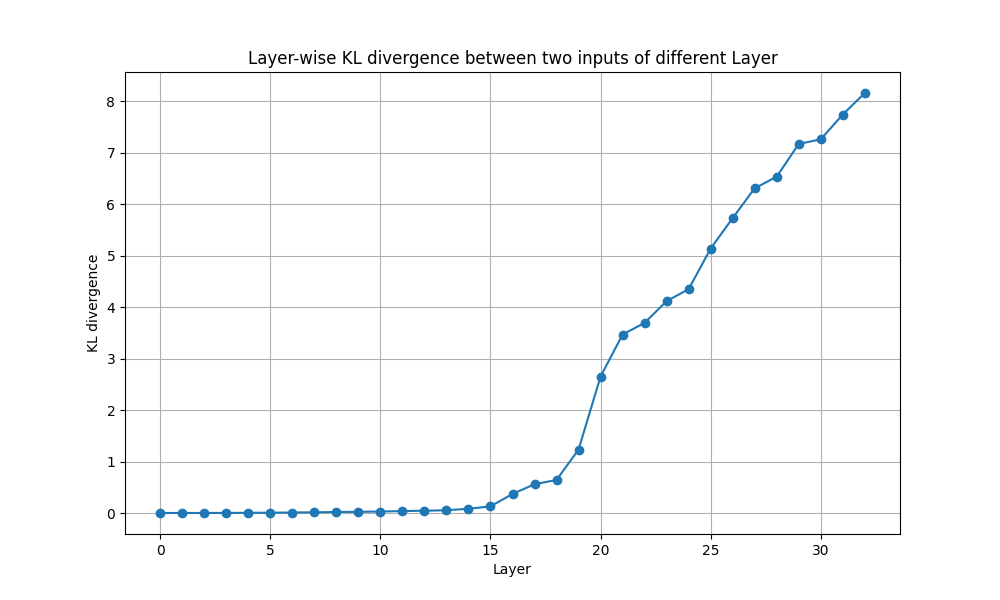}
    \caption{Hallucination}
    \label{fig:hallu}
    \end{subfigure}
    
    \caption{KL-Divergence Between Sampling Subspaces of different prompts Under Non-Hallucination and Hallucination Scenarios in LLMs.}
    \label{fig:KL}

\end{figure*}

Figure~\ref{fig:JSD} presents the Jensen-Shannon Divergence (JSD) between sampling subspaces across different layers of the LLM when prompted with semantically equivalent but syntactically different queries. 
The accompanying Table~\ref{tab:PromptAnswer} details the model's responses to each prompt. 
When the factual answer is known, the model provides consistent responses across prompts, resulting in a small JSD between subspaces. However, due to the cognitive uncertainty, variations in the prompts can lead to divergent responses, reflected in an increasing JSD with deeper layers. If the JSD exceeds a predefined threshold($\delta=1.0$), it indicates hallucination in the LLM's generation.

 In the early stage of exploration and experimentation, we compare two metrics-KL Divergence (KL-Divergence) and Jensen-Shannon Divergence (JSD)-in the hallucination detection module during the testing phase of our work. The choice of JSD is supported by both empirical observations and mathematical analysis, as detailed below.

\textbf{Empirical Observations.}
In our hallucination detection experiments, we measured the divergence of internal representations across different syntactic formulations of the same query using both JSD and KL-Divergence. The results reveal a key distinction:
    When hallucinations occur, JSD exhibits a sharp increase in intermediate layers (e.g., layers 15–25) and remains above a certain threshold, while it stabilizes below the threshold in non-hallucinatory scenarios (as shown is Figure~\ref{fig:JSD}).
   However, KL-Divergence shows elevated values in intermediate layers regardless of hallucination status, failing to differentiate hallucinatory outputs from confident predictions (as shown is Figure~\ref{fig:KL}).
   
   This divergence highlights fundamental differences in their mathematical properties and their applicability to uncertainty quantification.

\textbf{Mathematical Analysis of JSD and KL-Divergence.}
Let $P$ and $Q$ denote the probability distributions of model responses to two syntactically distinct but semantically equivalent prompts. The formal definition of KL-Divergence is as follows:
 \begin{equation}
     D_{\text{KL}}(P \parallel Q) = \sum_{i} P(i) \log \frac{P(i)}{Q(i)}   .
 \end{equation}
 \begin{enumerate}
     \item Asymmetry: $ D_{\text{KL}}(P \parallel Q) \neq D_{\text{KL}}(Q \parallel P) $
     \item Unbounded Range:$ D_{\text{KL}} \in [0, +\infty)$,  which is sensitive to outliers when $Q(i) \to 0$.
     \item Interpretation: Quantifies the information loss when approximating $P$ with $Q$.
 \end{enumerate}

The formal definition of JSD is as follows:
 \begin{equation}
     \text{JSD}(P \parallel Q) = \frac{1}{2} D_{\text{KL}}(P \parallel M) + \frac{1}{2} D_{\text{KL}}(Q \parallel M),
 \end{equation}
 where $ M = \frac{P + Q}{2} $.
 \begin{enumerate}
     \item Symmetry: $\text{JSD}(P \parallel Q) = \text{JSD}(Q \parallel P)$ 
     \item Bounded Range: $ \text{JSD} \in [0, 1]$, which is robust to distribution tails.
     \item Interpretation: Measures the average divergence of $P$ and $Q$ from their midpoint distribution $M$.
 \end{enumerate}

The suitability of JSD for hallucination detection arises from its symmetric and bounded nature, which aligns well with the requirements for detecting hallucinations caused by cognitive uncertainty: First, hallucinations manifest as inconsistencies in reasoning paths under varying syntactic contexts. JSD symmetrically captures the mutual disagreement between $P$ and $Q$, reflecting intrinsic uncertainty in the model's knowledge representation. 
 In contrast, KL-Divergence's asymmetry introduces bias: $D_{\text{KL}}(P \parallel Q)$ penalizes $Q$'s deviations from $P$ but not vice versa, amplifying noise in intermediate layers where syntactic variations naturally induce transient shifts in representation (e.g., rephrasing-triggered attention redistribution), which can lead to false positives.
 Second, intermediate layers often encode syntax-sensitive features (e.g., grammatical structures), which vary significantly across paraphrased queries even in non-hallucinatory scenarios. KL-Divergence's unbounded sensitivity to such variations causes spurious spikes, whereas JSD's boundedness filters out syntax-driven noise while retaining semantic-level inconsistencies.

The symmetry, boundedness, and robustness of JSD to syntactic noise make it particularly well-suited for detecting hallucinations driven by cognitive uncertainty. In contrast, the sensitivity of KL-Divergence to transient syntactic variations and directional bias limits its effectiveness in this context. Therefore, we have selected JSD as the metric for our unsupervised hallucination detection module.

\textbf{Hyperparameter analysis} 
To evaluate the robustness of our hallucination detection threshold $\delta$, we conduct a sensitivity analysis by varying $\delta$ from 0.6 to 1.5 in increments of 0.1, and measuring the resulting EM scores on three datasets: HotpotQA, 2Wiki, and TriviaQA. As shown in Table~\ref{tab:delta_sensitivity}, when the hallucination detection threshold is set too low, external knowledge is introduced without sufficient control, which may interfere with the LLM's reasoning process. As the threshold $\delta$ increases from 0.6 to 1.0, performance improves steadily, reaching its optimal at $\delta = 1.0$. Within the range $\delta \in [0.8, 1.2]$, EM scores remain stable with minimal fluctuation. However, when $\delta$ exceeds 1.2, the performance begins to degrade, likely due to the over-restriction of external knowledge retrieval. 

\begin{table}[ht]
\centering

\begin{tabular}{c|c|c|c}
\hline
 \multirow{2}*{\textbf{$\delta$}} & \textbf{HotpotQA} & \textbf{2Wiki} & \textbf{TriviaQA} \\
 \cmidrule(r){2-2} \cmidrule(r){3-3} \cmidrule(r){4-4}  
 \noalign{\vskip 0.15mm}
 &  \textbf{EM} & \textbf{EM} & \textbf{EM} \\
 
\hline
0.6 & 27.6 & 30.1 & 52.9 \\
0.7 & 27.9 & 30.4 & 53.3 \\
0.8 & 28.1 & 30.9 & 53.9 \\
0.9 & 28.4 & 31.4 & 54.3 \\
\textbf{1.0} & \textbf{28.5} & \textbf{31.6} & \textbf{54.5} \\
1.1 & 28.3 & 31.5 & 54.4 \\
1.2 & 27.9 & 31.0 & 54.0 \\
1.3 & 25.3 & 28.5 & 50.1 \\
1.4 & 23.1 & 25.3 & 47.6 \\
1.5 & 20.4 & 19.5 & 44.3 \\
\hline
\end{tabular}

\caption{Sensitivity analysis of detection threshold $\delta$ on EM scores.}
\label{tab:delta_sensitivity}
\end{table}

In the future work, we will introduce a more principled and adaptive mechanism for threshold selection to enhance the generality and flexibility of the framework. For example, potential future directions include: (1) dynamically adjusting $\delta$ based on statistical properties (e.g., the distribution of JSD across layers or samples), and (2) clustering-based threshold calibration. In the latter, we can apply unsupervised clustering algorithms (e.g., K-means or Gaussian Mixture Models) to the distribution of JSD values obtained from semantically perturbed prompts. By identifying natural groupings in the JSD space, for example, stable vs. unstable subspaces, the boundary between clusters can be used to automatically determine a context-dependent threshold $\delta$, eliminating the need for manual tuning and improving adaptability across models and domains.

\section{The details of knowledge filtering}
\label{sec:knowAppend}

\begin{figure*}
  \centering
    \includegraphics[width=\textwidth]{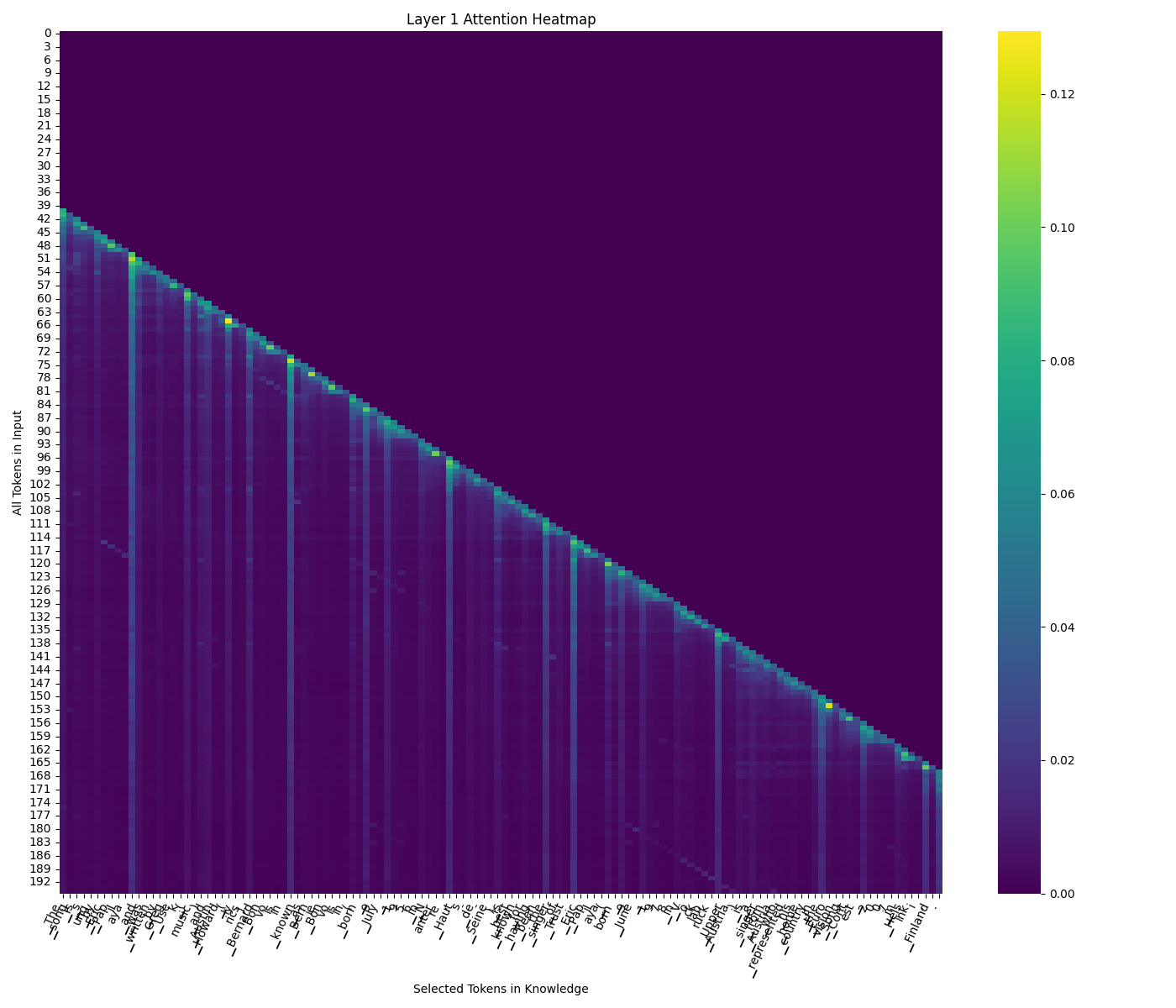}
    \caption{Attention heatmap of key layer}
    \label{fig:heatmapKey}

\end{figure*}

\begin{figure*}
  \centering
    \includegraphics[width=\textwidth]{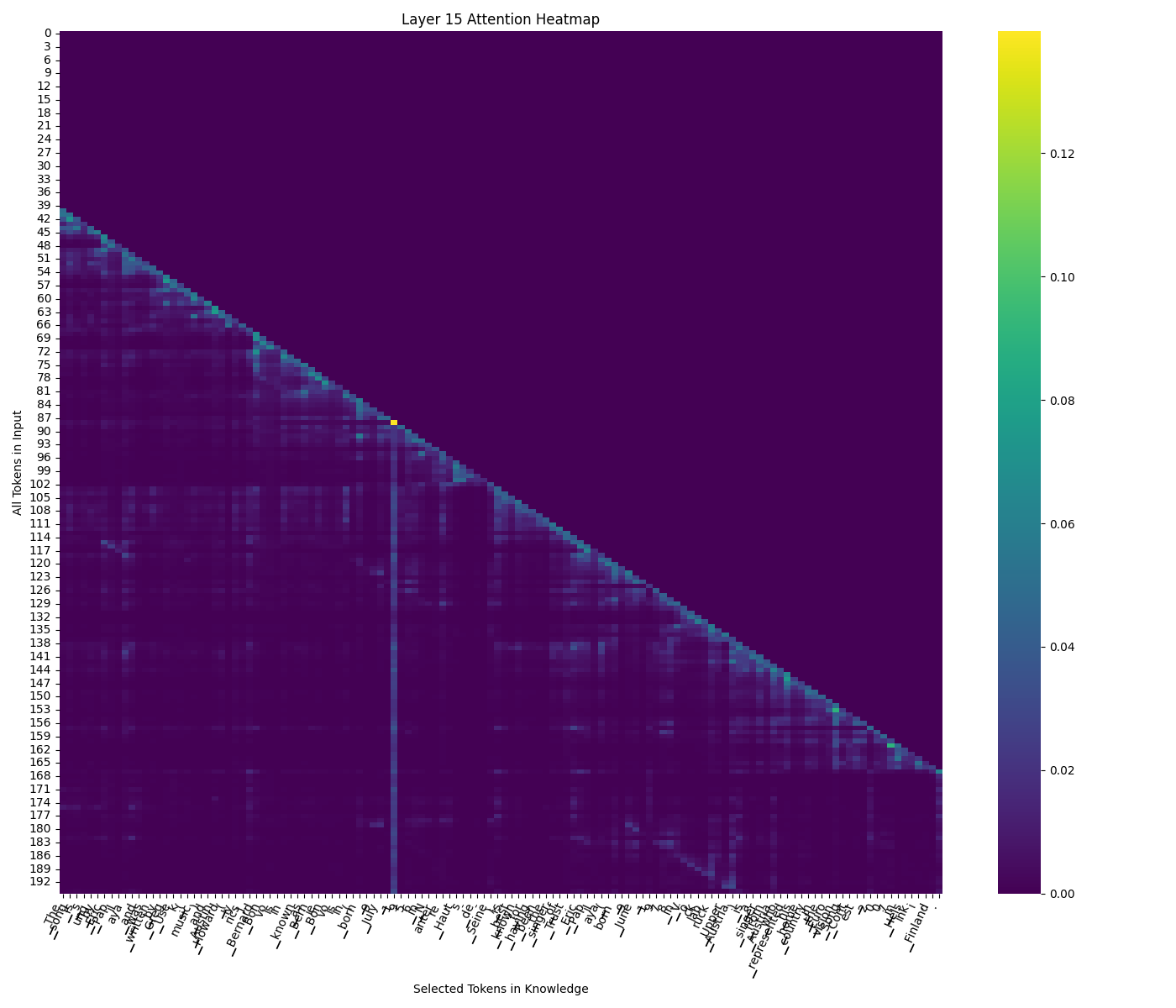}
    \caption{Attention heatmap of offset layer}
    \label{fig:heatmapOffset}

\end{figure*}

\begin{figure*}
  \centering
    \begin{subfigure}[b]{0.48\textwidth}
        \includegraphics[width=\textwidth]{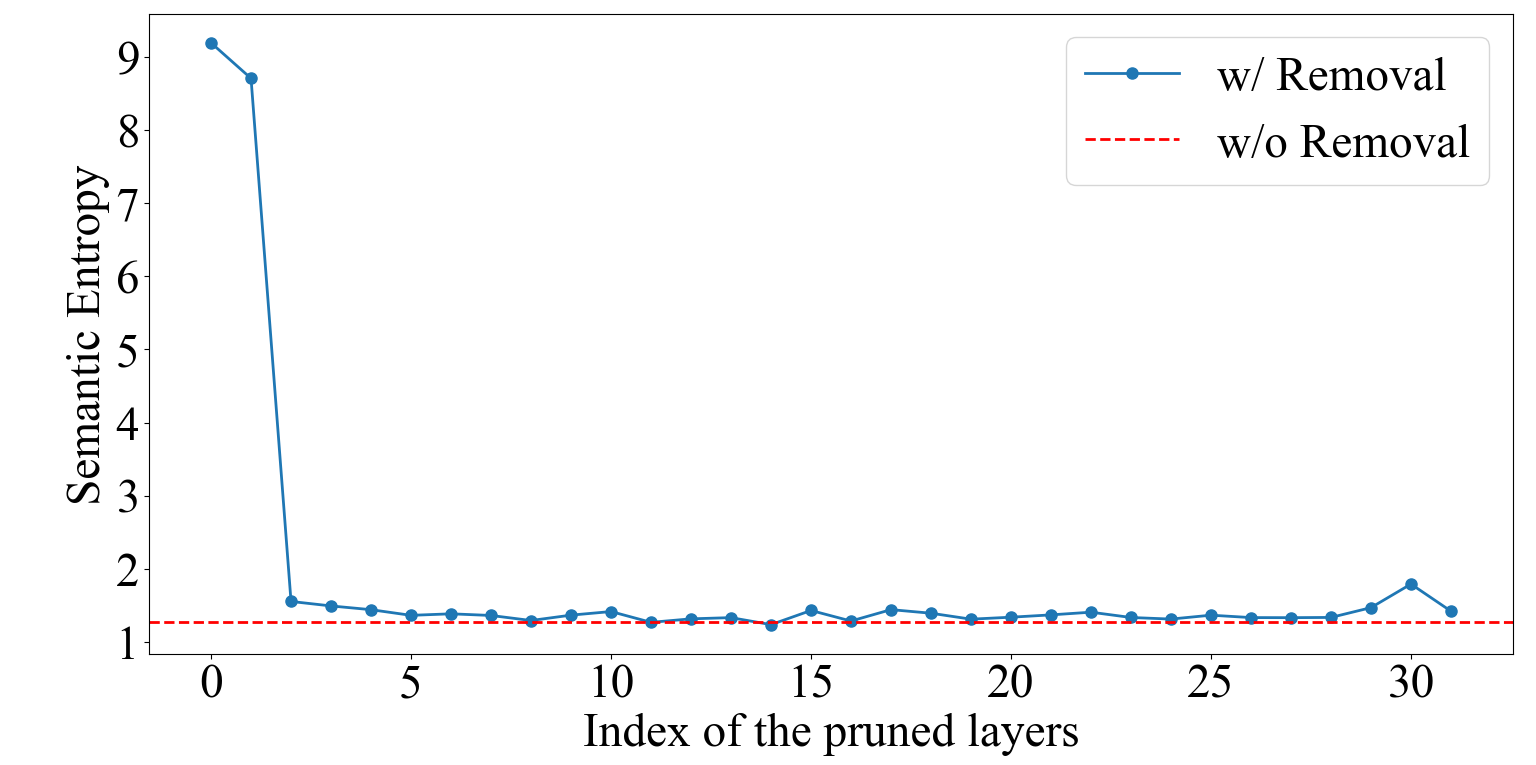} 
    \caption{Performance variation curves of the original models}
    \label{fig:curvesOriginal}
    \end{subfigure}
    \hfill
    \begin{subfigure}[b]{0.48\textwidth}
        \includegraphics[width=\textwidth]{figures/layer_importance_replot_with_attention.png}
    \caption{Performance variation curves of the fine-tuned models}
    \label{fig:curvesFinetune}
    \end{subfigure}
    
    \caption{Performance variation curves of the original and fine-tuned models under layer-wise pruning.}
    \label{fig:lossCurves}

\end{figure*}

We evaluate LLMs' attention distribution and response behavior when external knowledge is incorporated into the input query using the following prompts, where the red-highlighted segments indicate noise. As described in the methodology, the key layers in LLMs are responsible for capturing abstract global semantics, focusing on task-specific feature representation, and playing a central role in reasoning, as illustrated in Figure~\ref{fig:heatmapKey}. In contrast, offset layers tend to amplify noise, leading to deviations in attention distribution and adversely affecting prediction performance, as shown in Figure~\ref{fig:heatmapOffset}. Excessive reliance on external knowledge, coupled with noise interference, results in erroneous model outputs. Using attention distribution differences between the key and offset layers, we can effectively identify noise and mitigate its negative impact.
\begin{tcolorbox}[colback=gray!10,
                  colframe=black,
                  width=0.5\textwidth,
                  arc=2mm, 
                  auto outer arc,
                  title={Prompt of knowledge filtering},
                  breakable]
    \small
    \textit{User:} \\
    Please answer the given question with format referring to the knowledge provided below. \\
    \textbf{Question:} Where was the performer of song Get A Life – Get Alive born?\\
    \textbf{Knowledge:} The song is sung by Eric Papilaya, and was written by Greg Usek (music) and Austin Howard (lyrics). \textcolor{red}{Bernard Bonvoisin, known as Bernie Bonvoisin( born 9 July 1956 in Nanterre, Hauts- de- Seine), is best known for having been the singer of Trust.} Eric Papilaya (born 9 June 1978 in Vöcklabruck, Upper Austria) is a singer from Austria who represented his home country in the Eurovision Song Contest 2007 in Helsinki, Finland.\\
    \\
    \textit{Assistant:}\\
    \textbf{A: The answer is}\\

    \textit{Output:}
    \textcolor{red}{9 July 1956 in Nanterre, Hauts- de- Seine} 
    
\end{tcolorbox}
\vspace{-10pt}
\captionof{figure}{Prompt of knowledge filtering}  
\label{tcolorbox:KnowFilter} 

Figure~\ref{fig:lossCurves} presents the performance variation curves of the original and fine-tuned models under layer-wise pruning on the same test cases. The results indicate that semantic entropy, when computed based on the probability distribution of the original model, fails to serve as a reliable metric for evaluating key and offset layers. However, after fine-tuning with the proposed DSSP components and loss function, semantic entropy effectively captures the performance differences between the key and offset layer compared to the LLM without integrated knowledge.

Our approach leverages conditional entropy to mitigate uncertainty while using KL divergence to enhance distribution stability, thereby improving external knowledge filtering during the reasoning process. The fine-tuned model alleviates the inherent confidence bias of the original model, which may lead to excessive confidence in incorrect predictions or undue uncertainty in correct predictions.

\section{Mathematical derivation of the mixed attention for shared-private semantic decomposition}
\label{sec:mathDsspAppend}

In our work, to enhance generation process of LLM with the filtered external knowledge, DSSP module employs mixed attention to decompose dual-stream knowledge into shared and private semantics at the target layer. The semantic relevance between internal knowledge $X$ and external knowledge $Y$ is estimated based on dot product similarity, where the top $T$ tokens in the external knowledge that exhibit the highest similarity to the internal knowledge are selected as shared semantics.  Furthermore, we propose a differential attention mechanism to extract private semantics between internal and external knowledge. In the following, the detailed explanation and analysis of the underlying principles of differential attention are provided.

The differential attention mechanism is defined as the difference between self-attention $\tau (\cdot )$ of a single knowledge source and cross-attention $\eta (\cdot )$ of dual-stream knowledge:

\begin{equation}
  \mathcal{D}_{\text{attn}} (X,Y)=\tau (X,X)-\eta (X,Y)
\end{equation}

The mathematical explanation via explicit operator decomposition is as follows.

Let the latent representations of internal knowledge $X$ and external knowledge $Y$ be decomposed as follows:  

\begin{equation}
  X = S + P_X + N_X,
\end{equation}

\begin{equation}
  Y = S + P_Y + N_Y,
\end{equation} 

where $S$ represents the shared semantics (common component), $P_X, P_Y$ denote the private semantics of internal and external knowledge, respectively, and $N_X, N_Y$ correspond to noise components.

Traditional cross-attention $\eta(X, Y)$ primarily models the associations between $X$ and $Y$, and its output can be approximated as:

\begin{equation}
\begin{split}
    \eta(X, Y) &\propto \text{softmax} (X W_Q (Y W_K)^T ) Y W_V\\ &\approx \alpha S + \beta (N_X + N_Y),
\end{split}
\end{equation}

where $\alpha$ and $\beta$ are weighting coefficients. This formulation indicates that cross-attention predominantly captures a mixture of the shared semantics $S$ and noise components $N_X, N_Y$, while failing to effectively distinguish private semantics.

Conversely, self-attention $\tau(X, X)$ models the internal dependencies within $X$, and its output can be approximated as:

\begin{equation}
\begin{split}
    \tau(X, X) &\propto \text{softmax} (X W_Q (X W_K)^T ) X W_V\\ &\approx \gamma S + \delta P_X + \epsilon N_X ,
\end{split}
\end{equation}

which captures a combination of shared semantics $S$, private semantics $P_X$, and noise $N_X$.

By computing the difference between self-attention and cross-attention, $U_{\text{private}} \approx \tau(X, X) - \eta(X, Y)$,
the contribution of the shared semantics $S$ is effectively removed:

\begin{equation}
\begin{split}
    U_{\text{private}} &\approx (\gamma S + \delta P_X + \epsilon N_X) \\ &- (\alpha S + \beta (N_X + N_Y))  \\  
    &= (\gamma - \alpha) S + \delta P_X + (\epsilon - \beta) N_X - \beta N_Y ,
\end{split}
\end{equation}

By optimizing parameters such that $ \gamma \approx \alpha$ and $\epsilon \approx \beta$, the formulation is simplified as follows:

\begin{equation}
  U_{\text{private}} \approx \delta P_X - \beta N_Y.    
\end{equation}

The above derivation process demonstrates that the differential attention mechanism suppresses both shared semantics $S$ and internal noise $N_X$, while preserving private semantics $P_X$ and attenuating the impact of external noise $N_Y$.

\section{Prompt of query variation}
\label{sec:queryVaryAppend}

The prompt designed to enable the LLM to automatically generate alternative expressions, which are semantically similar but syntactically distinct from the original queries, is as follows:

\begin{tcolorbox}[colback=gray!10,
                  colframe=black,
                  width=0.5\textwidth,
                  arc=2mm, 
                  auto outer arc,
                  title={Prompt of query variation},
                  breakable]
    \small
    \textit{User:} \\
    Please provide another expression with the same meaning as the given sentence following the given example, and ensuring that both expressions have the same tokenization length using AutoTokenizer. \\
    \textbf{Sentence:} Who is the paternal grandfather of Tancred, King Of Sicily?
     \\
    \textbf{Another Sentence:} Tancred, King Of Sicily's paternal grandfather is who?\\
    \textbf{Sentence:} What is cause of death of performer of song How Do U Want It?
     \\
    \textbf{Another Sentence:} performer of song How Do U Want It, what is cause him die?\\
    \textbf{Sentence:} Where was the place of death of Yazid III's father?  \\
    \textbf{Another Sentence:} father of Yazid III Where was place of death of him?\\
    \textbf{Sentence:} [input]\\
    \textit{Assistant:}\\
    \textbf{Another Sentence:}
    
\end{tcolorbox}
\vspace{-13pt}
\captionof{figure}{Prompt of query variation}
\label{tcolorbox:queryVariation}

\section{Case study}
\label{sec:caseStudy}

To illustrate the practical implications of DSSP-RAG's Shared-Private Semantic Synergy in real-world tasks, we present representative examples in Table~\ref{tab:case_study_DSSP}. The first case demonstrates how the DSSP module effectively prevents the LLM from over-relying on noisy information introduced by the retrieved context during reasoning.
The second case illustrates that, under partial guidance from external knowledge, the DSSP module enables the LLM to successfully leverage its parametric knowledge to produce correct reasoning, thereby achieving a synergistic integration of internal and external knowledge. 

\begin{table*}[ht]
\centering
\small
\begin{tabularx}{\textwidth}{lX}
\toprule
\textbf{Question} & Where was the performer of song \textit{Get A Life – Get Alive} born? \\
\textbf{Ground Truth} & Vöcklabruck \\
\textbf{Context} & Bernard Bonvoisin, known as Bernie Bonvoisin (born 9 July 1956 in Nanterre, de-Seine), is best known for having been the singer of Trust \textbf{[Noise]}. The song is sung by Eric Papilaya, and was written by Greg Usek (music) and Austin Howard (lyrics). Eric Papilaya (born 9 June 1978 in Vöcklabruck, Upper Austria) is a singer from Austria who represented his home country in the Eurovision Song Contest 2007 in Helsinki, Finland. \\
\textbf{Output} & 
\begin{tabular}[t]{ll}
w/o DSSP & Nanterre, de-Seine \\
w/ DSSP & \textbf{Vöcklabruck} \\
\end{tabular} \\
\midrule
\textbf{Question} & Who is the paternal grandfather of Tancred, King Of Sicily? \\
\textbf{Ground Truth} & Roger II of Sicily \\
\textbf{Context} & Roger's first public act took place at Melfi in 1129, where, though still a child, he accepted the fealty of some rebellious barons along with his father and his younger brother Tancred \textbf{[Noise]}. Tancred was King of Sicily from 1189 to 1194. He was born in Lecce, an illegitimate son of Roger III, Duke of Apulia (the eldest son of King Roger II) by his mistress Emma, a daughter of Achard II, Count of Lecce. \\
\textbf{Output} & 
\begin{tabular}[t]{ll}
w/o DSSP & Roger I of Sicily \\
w/ DSSP & \textbf{Roger II of Sicily} \\
\end{tabular} \\
\bottomrule
\end{tabularx}
\caption{Case study: DSSP mitigates hallucinations via synergizing parametric and external knowledge under conflicting contexts.}
\label{tab:case_study_DSSP}
\end{table*}

\section{In context learning examples}
\label{sec:ICLAppend}

For DSSP\_SR, the retrieved external knowledge and query are provided separately as inputs to the LLM. At the target layer, the hybrid attention mechanism of the DSSP module integrates external knowledge with the parametric knowledge of the model to generate responses, as illustrated in Figure~\ref{fig:Framework}.  For DSSP\_adaptive, to maximally activate the LLM's parametric knowledge and reasoning capabilities, we follow the SEAKR approach by providing identical in-context learning examples for single-hop QA datasets, as shown in Figure~\ref{tcolorbox:singleICL}. For multi-hop QA datasets, we leverage IRCoT to supply dataset-specific in-context learning examples: HotpotQA (Figure~\ref{tcolorbox:HotpotICL}) and 2WikiMultiHopQA (Figure~\ref{tcolorbox:WikiICL}). The retrieved knowledge remains a separate input to the LLM, where it is first encoded and subsequently integrated with the parametric knowledge of the model through the hybrid attention mechanism of the DSSP module, enhancing both the generation process and the accuracy of the reasoning.

\begin{figure*}[t]  
\begin{tcolorbox}[colback=gray!10,
                  colframe=black,
                  title={single-hop QA}]
    \textbf{Question:} Nobody Loves You was written by John Lennon and released on what album that was issued by Apple
Records, and was written, recorded, and released during his 18 month separation from Yoko Ono? \\
\textbf{Answer:} The album issued by Apple Records, and written, recorded, and released during John Lennon's 18 month
separation from Yoko Ono is Walls and Bridges.
Nobody Loves You was written by John Lennon on Walls and Bridges album.
So the answer is Walls and Bridges. \\
\textbf{Question:} What is known as the Kingdom and has National Route 13 stretching towards its border? \\
\textbf{Answer:} Cambodia is officially known as the Kingdom of Cambodia.
National Route 13 stretches towards the border to Cambodia.
So the answer is Cambodia. \\
\textbf{Question:} Jeremy Theobald and Christopher Nolan share what profession? \\
\textbf{Answer:} Jeremy Theobald is an actor and producer.
Christopher Nolan is a director, producer, and screenwriter.
Therefore, they both share the profession of being a producer.
So the answer is producer. \\
\textbf{Question:} What film directed by Brian Patrick Butler was inspired by a film directed by F.W. Murnau? \\
\textbf{Answer:}  Brian Patrick Butler directed the film The Phantom Hour.
The Phantom Hour was inspired by the films such as Nosferatu and The Cabinet of Dr. Caligari.
Of these, Nosferatu was directed by F.W. Murnau.
So the answer is The Phantom Hour. \\
\textbf{Question:} Vertical Limit stars which actor who also played astronaut Alan Shepard in 'The Right Stuff'? \\
\textbf{Answer:} The actor who played astronaut Alan Shepard in 'The Right Stuff' is Scott Glenn.
The movie Vertical Limit also starred Scott Glenn.
So the answer is Scott Glenn. \\
\textbf{Question:} Which car, produced by Ferrari from 1962 to 1964 for homologation into the FIA's Group 3 Grand Touring
Car category inspired the Vandenbrink GTO? \\
\textbf{Answer:} The car produced by Ferrari from 1962 to 1964 for homologation into the FIA's Group 3 Grand Touring Car
category is the Ferrari 250 GTO.
The Ferrari 250 GTO also inspired the Vandenbrink GTO's styling.
So the answer is Ferrari 250 GTO.\\

Following the examples above, answer the question by reasoning step-by-step.\\
\textbf{Question:} [Question]

\end{tcolorbox}
\captionof{figure}{Examples for single-hop QA datasets}
\label{tcolorbox:singleICL}
\end{figure*}

\begin{figure*}[t]  
\begin{tcolorbox}[colback=gray!10,
                  colframe=black,
                  title={HotpotQA}]
    \textbf{Question:} Jeremy Theobald and Christopher Nolan share what profession?\\
\textbf{Answer:} Jeremy Theobald is an actor and producer.
Christopher Nolan is a director, producer, and screenwriter.
Therefore, they both share the profession of being a producer.
So the answer is producer.\\
\textbf{Question:} What film directed by Brian Patrick Butler was inspired by a film directed by F.W. Murnau?\\
\textbf{Answer:} Brian Patrick Butler directed the film The Phantom Hour.
The Phantom Hour was inspired by the films such as Nosferatu and The Cabinet of Dr. Caligari.
Of these, Nosferatu was directed by F.W. Murnau.
So the answer is The Phantom Hour.\\
\textbf{Question:} How many episodes were in the South Korean television series in which Ryu Hye-young played Bo-ra? \\
\textbf{Answer:} The South Korean television series in which Ryu Hye-young played Bo-ra is Reply 1988.
The number of episodes Reply 1988 has is 20.
So the answer is 20.\\
\textbf{Question:} Were Lonny and Allure both founded in the 1990s? \\
\textbf{Answer:} Lonny (magazine) was founded in 2009.
Allure (magazine) was founded in 1991.
Thus, of the two, only Allure was founded in the 1990s.
So the answer is no.\\
\textbf{Question:} Vertical Limit stars which actor who also played astronaut Alan Shepard in The Right Stuff ?\\
\textbf{Answer:} The actor who played astronaut Alan Shepard in The Right Stuff is Scott Glenn.
The movie Vertical Limit also starred Scott Glenn.
So the answer is Scott Glenn. \\
\textbf{Question:} What was the 2014 population of the city where Lake Wales Medical Center is located? \\
\textbf{Answer:} Lake Wales Medical Center is located in the city of Lake Wales, Polk County, Florida.
The population of Lake Wales in 2014 was 15,140.
So the answer is 15,140. \\
\textbf{Question:} Who was born first? Jan de Bont or Raoul Walsh? \\
\textbf{Answer:} Jan de Bont was born on 22 October 1943.
Raoul Walsh was born on March 11, 1887.
Thus, Raoul Walsh was born first.
So the answer is Raoul Walsh. \\
\textbf{Question:} In what country was Lost Gravity manufactured? \\
\textbf{Answer:} The Lost Gravity (roller coaster) was manufactured by Mack Rides.
Mack Rides is a German company.
So the answer is Germany.\\

Following the examples above, answer the question by reasoning step-by-step.\\
\textbf{Question:} [Question]

\end{tcolorbox}
\captionof{figure}{Examples for HotpotQA}
\label{tcolorbox:HotpotICL}
\end{figure*}

\begin{figure*}[t]  
\begin{tcolorbox}[colback=gray!10,
                  colframe=black,
                  title={2WikiMultihopQA}]
    \textbf{Question:} Who was born first out of Martin Hodge and Ivania Martinich?\\
\textbf{Answer:} Martin Hodge was born on 4 February 1959.
Ivania Martinich was born on 25 July 1995.
Thus, 4 February 1959 is earlier than 25 July 1995 and Martin Hodge was born first.
So the answer is Martin Hodge.\\
\textbf{Question:} When did the director of film Hypocrite (Film) die?\\
\textbf{Answer:} The film Hypocrite was directed by Miguel Morayta.
Miguel Morayta died on 19 June 2013.
So the answer is 19 June 2013.\\
\textbf{Question:} Are both Kurram Garhi and Trojkrsti located in the same country? \\
\textbf{Answer:} Kurram Garhi is located in the country of Pakistan.
Trojkrsti is located in the country of Republic of Macedonia.
Thus, they are not in the same country.
So the answer is no.\\
\textbf{Question:} Do the director of film Coolie No. 1 (1995 Film) and the director of film The Sensational Trial have the same
nationality? \\
\textbf{Answer:} Coolie No. 1 (1995 film) was directed by David Dhawan. \\
The Sensational Trial was directed by Karl Freund.
David Dhawan's nationality is Indian.
Karl Freund's nationality is German.
Thus, they do not have the same nationality.
So the answer is no.
\textbf{Question:} Who is Boraqchin (Wife Of Ögedei)'s father-in-law? \\
\textbf{Answer:} Boraqchin is married to Ögedei Khan.
Ögedei Khan's father is Genghis Khan.
Thus, Boraqchin's father-in-law is Genghis Khan.
So the answer is Genghis Khan.\\
\textbf{Question:} When did the director of film Laughter In Hell die? \\
\textbf{Answer:}The film Laughter In Hell was directed by Edward L. Cahn.
Edward L. Cahn died on August 25, 1963.
So the answer is August 25, 1963.\\

Following the examples above, answer the question by reasoning step-by-step.\\
\textbf{Question:} [Question]

\end{tcolorbox}
\captionof{figure}{Examples for 2WikiMultihopQA}
\label{tcolorbox:WikiICL}
\end{figure*}

\end{document}